\documentclass[lettersize,journal]{IEEEtran} 
\usepackage{array}
\usepackage{times}
\usepackage{textcomp} 
\usepackage[caption=false,font=normalsize,labelfont=sf,textfont=sf]{subfig}
\usepackage{stfloats}
\usepackage{verbatim}
\usepackage[numbers]{natbib}
\usepackage{cite}
\usepackage{microtype}
\usepackage{booktabs}
\usepackage{acronym}
\usepackage{amsfonts}
\usepackage{nicefrac}
\usepackage{amsmath}
\usepackage{amssymb}
\usepackage{amsthm}
\usepackage{amstext}
\usepackage{mathtools}
\usepackage[inline]{{enumitem}}
\usepackage{bbm}
\usepackage{algorithm}
\usepackage{algorithmic}
\usepackage{graphicx}
\usepackage{wrapfig}
\usepackage{multirow}

\usepackage{rotating}
\usepackage{latexsym}
\usepackage{hyperref}
\usepackage{cleveref}
\usepackage{fancyhdr}

\def\BibTeX{{\rm B\kern-.05em{\sc i\kern-.025em b}\kern-.08em                   
    T\kern-.1667em\lower.7ex\hbox{E}\kern-.125emX}}                             
\usepackage{balance}

\newtheorem{theorem}{Theorem}

\newtheorem{lemma}[theorem]{Lemma}

\usepackage{amsmath,amsfonts,bm}

\def\eqref#1{equation~\ref{#1}}

\def\1{\bm{1}}

\DeclareMathAlphabet{\mathsfit}{\encodingdefault}{\sfdefault}{m}{sl}
\SetMathAlphabet{\mathsfit}{bold}{\encodingdefault}{\sfdefault}{bx}{n}

\newcommand{\Prob}{\mathrm{Pr}}

\DeclareMathOperator*{\argmin}{arg\,min}

\newtheorem{innercustomgeneric}{\customgenericname}
\providecommand{\customgenericname}{}
\newcommand{\newcustomtheorem}[2]{%
  \newenvironment{#1}[1]
  {%
   \renewcommand\customgenericname{#2}%
   \renewcommand\theinnercustomgeneric{##1}%
   \innercustomgeneric
  }
  {\endinnercustomgeneric}
}

\newcustomtheorem{customthm}{Theorem}
\newcustomtheorem{customcorl}{Corollary}
\newcustomtheorem{customfct}{Fact}
\newcustomtheorem{customlemma}{Lemma}

\fancypagestyle{firstpage}{%
  \renewcommand{\headrulewidth}{0pt}
  \lhead{}
  \lfoot{\scriptsize © 2021 IEEE. Personal use of this material is permitted. Permission
from IEEE must be obtained for all other uses, in any current or future
media, including reprinting/republishing this material for advertising or
promotional purposes, creating new collective works, for resale or
redistribution to servers or lists, or reuse of any copyrighted
component of this work in other works.}
}

\acrodef{MER}{Meta-Experience Replay}
\acrodef{GEM}{Gradient Episodic Memory}
\acrodef{A-GEM}{Averaged-\ac{GEM}}
\acrodef{OGD}{Orthogonal Gradient Descent}
\acrodef{GSS}{Gradient-based Sample Selection}
\acrodef{SDRL}{Semi-Discriminative Representation Loss}
\acrodef{ER}{Experience Replay}
\acrodef{BER}{Balanced Experience Replay}
\acrodef{GSS-IQP}{\acl{GSS} with Interger Quadratic Programming}
\acrodef{GSNR}{Gradient Signal to Noise Ratio}
\acrodef{SNR}{Signal to Noise Ratio}
\acrodef{SGD}{Stochastic Gradient Descent}
\acrodef{KFD}{Kernel Fisher Discriminant analysis}
\acrodef{DML}{Deep Metric Learning}
\acrodef{SOTA}{state-of-the-art}
\acrodef{MLP}{Multi-layer Perceptron}

\newcommand{\cl}{{{continual learning}}}

\Crefname{equation}{Eq.}{Eqs.}
\Crefname{figure}{Fig.}{Figs.}
\Crefname{section}{Sec.}{Secs.}
\Crefname{algorithm}{Alg.}{Algs.}
\Crefname{theorem}{Theorem}{Theorems}
\Crefname{appendix}{Appx.}{Appxs.}
\Crefname{table}{Tab.}{Tabs.}

\begin{document}

\title{\acl{SDRL} \\ for Online Continual Learning }

\author{Yu Chen, 
        Tom Diethe,        
        Peter Flach  
        \thanks{Yu Chen and Peter Flach are with Department of Computer Science, University of Bristol, Bristol, UK. (e-mail: yc14600@bristol.ac.uk)}
        \thanks{Tom Diethe is with Amazon Research, Seattle, US.}
}
%
\markboth{In submission}
{Yu Chen,
\MakeLowercase{\textit{(et al.)}}: \acl{SDRL} for Online Continual Learning}

\maketitle
\thispagestyle{firstpage}
\begin{abstract}
The use of episodic memory in continual learning has demonstrated effectiveness for alleviating catastrophic forgetting. In recent studies, gradient-based approaches have been developed to make more efficient use of compact episodic memory. Such approaches refine the gradients resulting from new samples by those from memorized samples, aiming to reduce the diversity of gradients from different tasks. 
In this paper, we clarify the relation between diversity of gradients and discriminativeness of representations, showing shared as well as conflicting interests between \acl{DML} and \cl{}, thus demonstrating pros and cons of learning discriminative representations in \cl{}. Based on these findings, we propose a simple method -- \acf{SDRL} -- for \cl{}. In comparison with state-of-the-art methods, \ac{SDRL} shows better performance with low computational cost on multiple benchmark tasks in the setting of online \cl{}. \end{abstract} 

\begin{IEEEkeywords}        
Online continual learning, \acl{GEM}, \acl{DML}, discriminative representation.                  
\end{IEEEkeywords}

\section{Introduction}\label{sec:intro}
\IEEEPARstart{I}{n} the real world, we are often faced with situations where data distributions are changing over time, and we would like to update our models by new data in time, with bounded growth in system size and computation time. These situations fall under the umbrella of ``\cl{}'', which has many practical applications, such as recommender systems, retail supply chain optimization, and robotics \citep{lesort2019continual,diethe2018continual,tian2018continuum}. Comparisons have also been made with the way that humans are able to learn new tasks without forgetting previously learned ones, using common knowledge shared across different skills.
The fundamental problem in \cl{} is \emph{catastrophic forgetting} \citep{mccloskey1989catastrophic,kirkpatrick2017overcoming}, \emph{i.e.} machine learning models have a tendency to forget previously learned tasks while being trained on new ones.

Here we formalize the problem setting of \cl{} as follows. Suppose a model $f(\cdot; \bm{\theta})$ is a function mapping an input space to an output space and $\bm{\theta}$ represents the model parameters. It receives training datasets of a series of tasks sequentially. Let $\mathcal{D}_t = \{X_t,Y_t\}$ denote the training data of the $t$-th task, and $\bm{\theta}_t$ denotes model parameters at the $t$-th task. The optimization objective of \cl{} methods can be written as:
\begin{equation}
    \begin{split}
        \bm{{\theta}}^{*}_t = \argmin_{\bm{\theta}_t} \mathcal{L}(f(\{{X}_t, \tilde{X}_t\}; \{\bm{\theta}_t, \bm{{\theta}}^{*}_{t-1}\}),\{{Y}_t, \tilde{Y}_t\})
    \end{split}
\end{equation}
where $\mathcal{L}(\cdot)$ is the loss function, ${\mathcal{M}}_t = \{ \tilde{X}_t, \tilde{Y}_t \}$ represents data in  the episodic memory at the $t$-th task (which could store very limited samples from previous tasks). The goal is to obtain a model $f(\cdot;\bm{\theta}_t^{*})$ that can perform well on test data from all learned tasks. The difficulty stems from the limited resource (i.e. ${\mathcal{M}}_t, \bm{{\theta}}^{*}_{t-1}$) that is available for preserving information of previous tasks.

There are three main categories of methods in \cl{}:
\begin{enumerate*}[label=\emph{\roman*})]
    \item regularization-based methods which aim to preserve important parameters of the model trained upon previous tasks \citep{kirkpatrick2017overcoming,zenke2017continual,nguyen2017variational,ebrahimi2019uncertainty}; 
    \item architecture-based methods for incrementally evolving the model by learning task-shared and task-specific components \citep{schwarz2018progress,hung2019compacting,yoon2020scalable,zhang2020regularize};
    \item replay-based methods which focus on preserving the knowledge of previous data distributions by replaying data samples stored in the episodic memory or generated by a generative model   \citep{shin2017continual,rolnick2019experience,aljundi2019gradient,lopez2017gradient,chaudhry2018efficient}.
\end{enumerate*}

In general, replay-based methods with episodic memories are more  efficient than methods in other categories because this type of approach can provide competitive performance with much less computational cost. Particularly, replay-based approaches are more robust with more difficult settings in continual learning \citep{chaudhry2019continual,chrysakis2020online,prabhu2020gdumb}, such as online training (training with one epoch), task agnostic setting (no task identifier provided during testing time), task boundary agnostic setting (no clear task boundaries during training).  
The replay-based methods mostly attempt to efficiently utilize samples from the memories, including different sampling strategies \citep{aljundi2019online,chaudhry2019continual,chrysakis2020online}, methods of better knowledge transferring using the memorized samples \citep{rebuffi2017icarl,buzzega2020dark}, and methods of refining gradients from new samples by memorized samples \citep{lopez2017gradient,chaudhry2018efficient,riemer2018learning,farajtabar2020orthogonal}.

In particular, gradient-based approaches using episodic memories have been receiving increasing attention \citep{lopez2017gradient,chaudhry2018efficient,aljundi2019gradient,farajtabar2020orthogonal,guo2020improved,chaudhry2020continual}.
The essential idea is to reduce the performance degradation on old tasks by forcing the inner product of the two gradients to be non-negative \citep{lopez2017gradient}:
\begin{equation}\label{eq:gem_base}
    \langle \bm{g}_t,\bm{g}_k \rangle
    = \left \langle \frac{\partial\mathcal{L}(\mathbf{x}_t,\bm{\theta})}{\partial \bm{\theta}}, \frac{\partial\mathcal{L}(\mathbf{x}_k,\bm{\theta})}{\partial \bm{\theta}} \right\rangle
    \ge 0, \ \ \forall k < t
\end{equation}
where $t$ and $k$ are time indices, $\mathbf{x}_t$ denotes a new sample from the current task, and $\mathbf{x}_k$ denotes a sample from the episodic memory. Thus, the updates of parameters are forced to preserve the performance on previous tasks as much as possible. This idea indicates the samples' gradients with most negative inner products are critical to generalization across tasks. However, 
\citet{chaudhry2019continual} show that the average gradient over a small set of random samples may be able to obtain good generalization as well. 

In this paper, we try to answer the following questions:
\begin{enumerate}[label=\emph{\roman*})]
    \item Which samples tend to produce gradients that strongly diverse with other samples and why are such samples able to help with generalization? 
    \item Can we reduce diverse  gradients in a more efficient way? 
    \item Is reducing diverse gradients enough for solving problems of continual learning?
\end{enumerate}
Our answers reveal: 
\begin{enumerate*}[label=\emph{\arabic*})]
    \item  the relation between diverse gradients and discriminative representations, which connects \acf{DML} \citep{kaya2019deep,roth2020revisiting} and \cl{};
    \item pros and cons of learning discriminative representations in \cl{}.
\end{enumerate*}
Drawing on these findings we propose a new approach, \acf{SDRL}, for classification tasks in continual learning. 
In addition, we suggest a simple replay strategy \acf{BER} that  naturally complements \ac{SDRL} in the setting of online \cl{}. 
Comparing with several gradient-based methods, 
our methods show improved performance with much lower computational cost across multiple benchmark tasks and datasets.

\section{Related Work}\label{sec:related}

Ideally, every new sample should satisfy the inequality in \Cref{eq:gem_base}, however, it is not a straightforward objective in practice. In \ac{GEM} \citep{lopez2017gradient}, $g_t$ is projected to a new direction that is closest to itself in $L_2$-norm whilst satisfying \Cref{eq:gem_base}:
\begin{equation}\label{eq:gem}
     \bm{g}^{*} = \argmin_{\tilde{\bm{g}}} \frac{1}{2} ||\bm{g}_t - \tilde{\bm{g}}||_2^2, \ \ 
        s.t. \langle \tilde{\bm{g}}, \bm{g}_k \rangle \ge 0, \ \ \forall k < t
\end{equation}
Optimization of this objective requires a high-dimensional quadratic program and thus is computationally expensive. \ac{A-GEM} \citep{chaudhry2018efficient} alleviates the computational burden of \ac{GEM} by using the averaged gradient over a batch of samples instead of individual gradients of samples in the episodic memory. 
\begin{equation}
    \begin{split}
        &\bm{g}^{*} = \argmin_{\tilde{\bm{g}}} \frac{1}{2} ||\bm{g}_t - \tilde{\bm{g}}||_2^2, \ \ s.t. \ \  \langle \tilde{\bm{g}}, \bm{g}_{ref} \rangle \ge 0, 
    \end{split}
\end{equation}
where $\bm{g}_{ref}$ is the average gradients produced by a batch of memorized samples. It can be solved by $\tilde{\bm{g}} = \bm{g}_t - \frac{\bm{g}_t^T\bm{g}_{ref}}{\bm{g}_{ref}^T\bm{g}_{ref}}$ which only involves computing inner products of gradients.
This not only simplifies the computation, but  also obtains comparable   performance with \ac{GEM}. \citet{guo2020improved} proposed adaptive schemes based on \ac{A-GEM} to obtain adaptive weights on $\bm{g}_t$ and $\bm{g}_{ref}$ for a better performance.

\ac{OGD} \citep{farajtabar2020orthogonal} projects $\bm{g}_t$ to the direction that is perpendicular to the surface formed by $\{\bm{g}_k| k < t \}$, which needs to store all the gradients $\{\bm{g}_k| k < t \}$ and hence is very costly in terms of memory. \citet{chaudhry2020continual} proposed ORTHOG-SUBSPACE for learning orthonomal weight matrices between tasks so that the gradients of current and previous tasks are orthogonal. It first constructs a projection matrix $P_t$ for each task that satisfies 
\begin{equation}
    P_t^T P_t = I, \ \ P_t^T P_k = 0, \ \ \forall k \neq t,
\end{equation}
where $t$ and $k$ are task indices; and then projects the output of the final hidden layer (the representation to the linear layer) as $\bm{\phi}_t = P_t\bm{h}^{(L)}$. This projection guarantees $\langle \bm{g}^{(L)}_t, \bm{g}^{(L)}_{k\neq t} \rangle = 0$, i.e. $\bm{g}^{(L)}_t \perp \bm{g}^{(L)}_{k\neq t}$. To preserve the inner product of gradients in each layer, the model requires orthonormal weight matrices and hence the objective becomes an optimization problem over Stiefel Manifold \citep{bonnabel2013stochastic} and can be solved iteratively using the Cayley transform \citep{li2020efficient}. This method achieves a similar goal as \ac{OGD} without storing gradients of previous tasks. Nevertheless, it requires task-specific projection $P_t$ during testing which is infeasible in task agnostic scenarios.

Moreover, \citet{aljundi2019gradient} propose \ac{GSS}, which selects samples that produce most diverse gradients with other samples into the episodic memory. 
\begin{equation}\label{eq:GSS}
    \begin{split}
        \widehat{\mathcal{M}}_t &= \argmin_{\mathcal{M}} \sum_{n,m \in \mathcal{M}} \frac{\langle \bm{g}_n,\bm{g}_m \rangle}{||\bm{g}_n|| \cdot ||\bm{g}_m||}, \\ 
        &s.t. \ \ {\mathcal{M}} \subset (\widehat{\mathcal{M}}_{t-1} \cup \mathcal{D}_t), \ \ |{\mathcal{M}}| = M
    \end{split}
\end{equation}
Here $M$ is the memory size, $\mathcal{D}_t$ is the training data of the current task. 
Since the diversity is measured by the cosine similarity between gradients and the cosine similarity is computed using the inner product of two normalized gradients, \ac{GSS} embodies the same principle as other gradient-based approaches introduced above. \ac{GSS} tries to preserve more information of old task by storing most critical samples of those tasks.
However, due to the limited size of the episodic memory, \ac{GSS} faces difficulties to construct the memory that is representative enough for all seen tasks. 

One general issue of these gradient-based approaches is that the computational complexity in addition to common back-propagation is proportional to the number of model parameters, which causes much higher cost on larger models. We will show experimental results regarding this issue in \Cref{sec:experiments}.


\section{The source of gradient diversity}
\begin{figure}[t!]
    \centering

    \includegraphics[width=0.8\linewidth,trim={0.cm .1cm .7cm .8cm},clip]{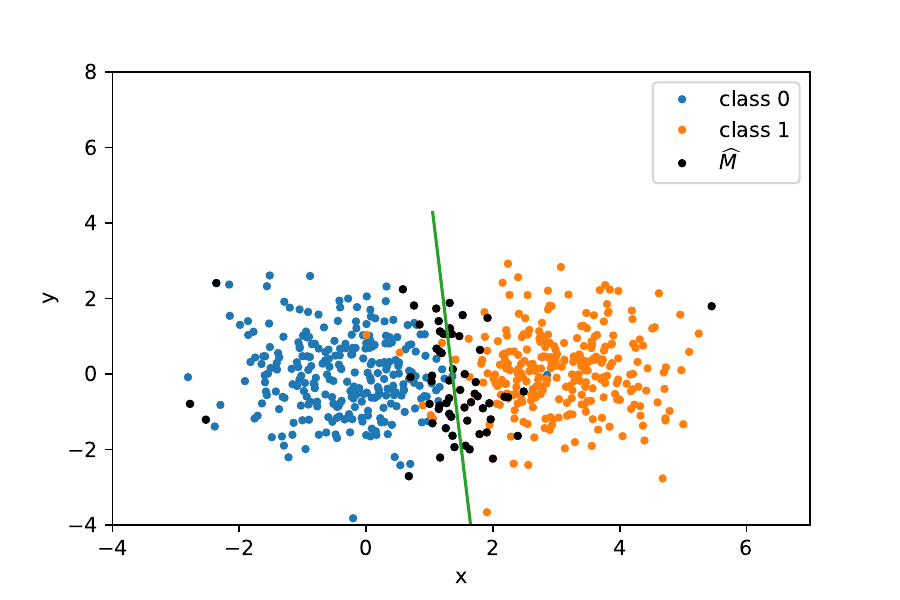}

    \caption{2-D classification examples, the $x$ and $y$ axis are the coordinates (also features) of samples. Samples with most diverse gradients ($\widehat{M}$) 
    are shown by black dots, the green line is the decision boundary.}
    \label{fig:2dim_example}
\vspace{-0.2cm}
\end{figure}

The common idea of gradient-based approaches indicates that samples with the most diverse gradients (the gradients have largely negative similarities with other samples) are most critical to generalization across tasks.
In this sense, we first tried to identify samples with most diverse gradients in a 2-D feature space by a simple  classification task of 2-D Gaussian distributions. 
We trained a linear model to discriminate between two classes (blue and orange dots in \Cref{fig:2dim_example}). We then select a set of samples $\widehat{\mathcal{M}}$ that produce gradients having the largest diversity (smallest similarity) with other samples as defined in \Cref{eq:GSS} (black dots in \Cref{fig:2dim_example}). The size of $\widehat{\mathcal{M}}$ is 10\% of the training set. 
%
%
It is clear from \Cref{fig:2dim_example} that the samples in $\widehat{\mathcal{M}}$ are mostly around the decision boundary between the two classes.
It matches the intuition  that gradients can be strongly opposed when samples from different classes are very similar. 
An extreme example would be two samples with same representations having different labels. In this case, the gradient to decrease the loss of one sample must increase the loss of the other.
The experimental results indicate that \textbf{\emph{more similar representations in different classes result in more diverse gradients.}} In the following we theoretically analyze this connection between the gradients and representations of a linear model.

\textbf{Notations}: \textbf{\emph{Negative pair}} means two samples from different classes; \textbf{\emph{positive pair}} means two samples from a same class. Let $\mathcal{L}$ represent the softmax cross entropy loss, $\mathbf{W} \in \mathbb{R}^{D \times K} $ is the weight matrix of the linear model, and $\mathbf{x}_n \in \mathbb{R}^{D}$ denotes $n$-th input data sample,  $\mathbf{y}_n \in \mathbb{R}^{K}$ is a one-hot vector that denotes the label of $\mathbf{x}_n$, $D$ is the dimension of data representations, 
$K$ is the number of classes. Let $\bm{p}_n = softmax(\mathbf{o}_n)$, where $\mathbf{o}_n = \mathbf{W}^T\mathbf{x}_n$, the gradient $\bm{g}_n = \nabla_{\mathbf{W}} \mathcal{L}(\mathbf{x}_n, \mathbf{y}_n;\mathbf{W})$. $\mathbf{x}_n,\mathbf{x}_m$ are two different samples when $n \neq m$. 
\begin{lemma}\label{lemma1}
Let $\bm{\epsilon}_n = \bm{p}_{n} - \mathbf{y}_{n}$, we have:
$
       \langle \bm{g}_{n}, \bm{g}_{m} \rangle = \langle \mathbf{x}_{n}, \mathbf{x}_{m} \rangle \langle \bm{\epsilon}_n, \bm{\epsilon}_m \rangle,
$
\end{lemma}

\begin{customthm}{1}\label{thm1}
    Suppose $\mathbf{y}_n \neq \mathbf{y}_m$, and let $c_n$ denote the class index of $\mathbf{x}_n$ (i.e. $\mathbf{y}_{n,c_n} = 1, \mathbf{y}_{n,i}=0, \forall i \neq c_n$). Let $\beta \triangleq \bm{p}_{n,c_m} + \bm{p}_{m,c_n}$ and  $s_p \triangleq \langle \bm{p}_n, \bm{p}_m \rangle$, then:
\begin{equation*}
    \begin{split}
       & \Prob\Bigg(\text{sign}(\langle \bm{g}_{n}, \bm{g}_{m} \rangle)= \text{sign}(-\langle \mathbf{x}_{n}, \mathbf{x}_{m} \rangle)\Bigg) 
        = \Prob(\beta > s_p),
    \end{split}
\end{equation*}
\end{customthm}
\begin{theorem}\label{thm2}
    Suppose $\mathbf{y}_n = \mathbf{y}_m$, when $\langle \bm{g}_{n}, \bm{g}_{m} \rangle \neq 0$, we have:
    \begin{equation*}
    \text{sign}(\langle \bm{g}_{n}, \bm{g}_{m} \rangle)= \text{sign}(\langle \mathbf{x}_{n}, \mathbf{x}_{m} \rangle)
    \end{equation*}
    \end{theorem}
Proof of the theorems can be found in Appendix A. Lemma 1 indicates that the absolute value of $\langle \bm{g_n}, \bm{g_m} \rangle$ largely depends on $\langle \mathbf{x}_n, \mathbf{x}_m \rangle$ when $\langle \bm{\epsilon}_n, \bm{\epsilon}_m \rangle$ is not close to zero  since $ \langle \bm{\epsilon}_n, \bm{\epsilon}_m \rangle$ is bounded.
Theorem 1 says that for samples from different classes, $\langle \bm{g_n}, \bm{g_m} \rangle$ gets an \emph{opposite} sign of $\langle \mathbf{x}_n, \mathbf{x}_m \rangle$ when $\beta  > s_p$, where $\beta$ reflects how likely the model may misclassify both samples to its opposite class and $s_p$ reflects the similarity of the two samples that is approximated by the model. Theorem 2 says that for samples from a same class $\langle \bm{g}_{n}, \bm{g}_{m} \rangle$ has the \emph{same} sign as $\langle \mathbf{x}_{n}, \mathbf{x}_{m} \rangle$.

\begin{figure}[t!]
    \centering

        \includegraphics[width=0.8\linewidth,trim={0.cm .1cm .7cm .8cm},clip]{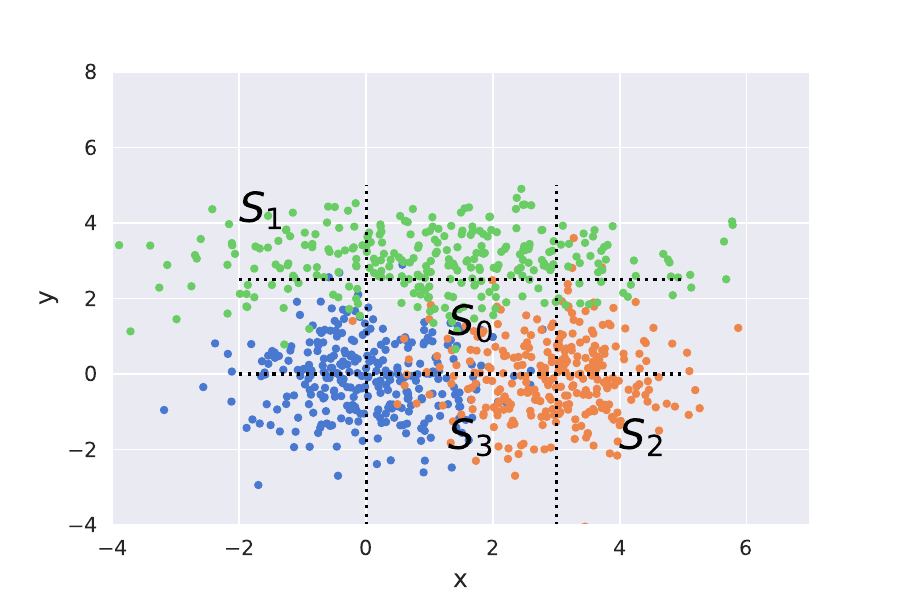}

    \caption{Splitting samples into several subsets in a 3-class classification task. Dots in different colors are from different classes. } 
\label{fig:4c_sets}
\end{figure}

\setlength{\tabcolsep}{3pt}
\begin{table*}[t!]
    \centering
    \caption{Illustration of the Theorems by drawing pairs of samples from different subsets that are defined in \Cref{fig:4c_sets}. 
    }
    \small
    \begin{tabular}{c|c|cccc|cccc}
    \hline
          &          & \multicolumn{4}{c|}{Negative pairs (Thm. 1)}            & \multicolumn{4}{c}{Positive pairs (Thm.2)}             \\ 
          \hline
                &     & $S_0$ & $S_0 \cup S_1$ & $S_3$ & $S_1 \cup S_2$ & $S_0$ & $S_0 \cup S_1$ & $S_3$ & $S_1 \cup S_2$ \\ 
                \hline
     & $\rm{Pr}(\langle \mathbf{x}_n, \mathbf{x}_m \rangle > 0)$ & 1.    & 0.877          & 1.    & 0.            & 1.    & 0.99          & 1.    & 1.             \\ 
     \hline
    \multirow{2}{*}{Linear} & $\rm{Pr}$$(\langle \bm{g}_n, \bm{g}_m \rangle < 0)$ & 0.727 & 0.725          & 1.    & 0.978          & 0.    & 0.007          & 0.    & 0.             \\ 
    
    & $\rm{Pr}$$(\beta  > s_p)$ & 0.727  & 0.687  & 1.  &  0. &  --  & -- & --  & -- \\
     \hline
    \multirow{2}{*}{MLP (ReLU)} & $\rm{Pr}$$(\langle \bm{g}_n, \bm{g}_m \rangle < 0)$ & 0.72 & 0.699          & 1.    & 0.21          & 0.013    & 0.01          & 0.    & 0.             \\ 
    
    & $\rm{Pr}(\beta  > s_p)$ & 0.746  & 0.701  & 1.  &  0. &  --  & -- & --  & -- \\ \hline
    \multirow{2}{*}{MLP (tanh)} & $\rm{Pr}$$(\langle \bm{g}_n, \bm{g}_m \rangle < 0)$ & 0.745 & 0.744          & 1.    & 0.993          & 0.004    & 0.007          & 0.    & 0.             \\ 
    & $\rm{Pr}(\beta  > s_p)$ & 0.766  & 0.734  & 1.  &  0. &  --  & -- & --  & -- \\ \hline
    \end{tabular}
    \label{tab:4c_prop}
\end{table*}

For a better understanding of the theorems, we conduct an empirical study by partitioning the feature space of three classes into several subsets as shown in \Cref{fig:4c_sets} and examine four cases of pairwise samples from these subsets: 1). $S_0$, both samples in a pair are near the intersection of all three classes; 2). $S_0 \cup S_1$, one sample is close to decision boundaries and the other is far away from the boundaries; 3). $S_3$, both samples close to the decision boundary between their true classes but away from the third class; 4). $S_1 \cup S_2$, both samples are far away from the decision boundaries. 

By training different models (a linear model, a \ac{MLP} with ReLU activation, and a \ac{MLP} with tanh activation) to classify the three classes shown in \Cref{fig:4c_sets}, we provide the empirical probability of getting negative inner product of gradients in \Cref{tab:4c_prop}. The gradients are computed by all parameters of the model. We can see that the non-linear models (MLPs) exhibit similar behaviors with the linear model, which are highly consistent with the theorems. For example, when most $\langle \mathbf{x}_n, \mathbf{x}_m \rangle$ are positive and  $\text{Pr}(\beta  > s_p)$ is high,   $\text{Pr}(\langle \bm{g}_n, \bm{g}_m \rangle < 0)$ is also high for negative pairs. 
One exception is that the \acs{MLP} with ReLU gets much less negative $\langle \bm{g}_n, \bm{g}_m \rangle$ by negative pairs in the case of $S_1 \cup S_2$, we consider the difference is caused by representations to the final linear layer always being positive due to ReLU activations. One evidence is that the \ac{MLP} with tanh activation still aligns with the linear model in this case. It is worth to note that the results of positive pairs in \Cref{tab:4c_prop} match Theorem 2 very well for all models. 
As $\langle \mathbf{x}_n, \mathbf{x}_m \rangle$ is mostly positive for positive pairs, $\langle \bm{g}_n, \bm{g}_m \rangle$ hence is also mostly positive. These results indicate that: \emph{when representations are non-negative, gradients having negative inner products are mostly from negative pairs and highly correlate to $\text{Pr}(\beta  > s_p)$}. 

\begin{figure*}[t!]
    \centering
\subfloat[]{
        \includegraphics[width=0.45\linewidth,trim={0.2cm .1cm .7cm .8cm},clip]{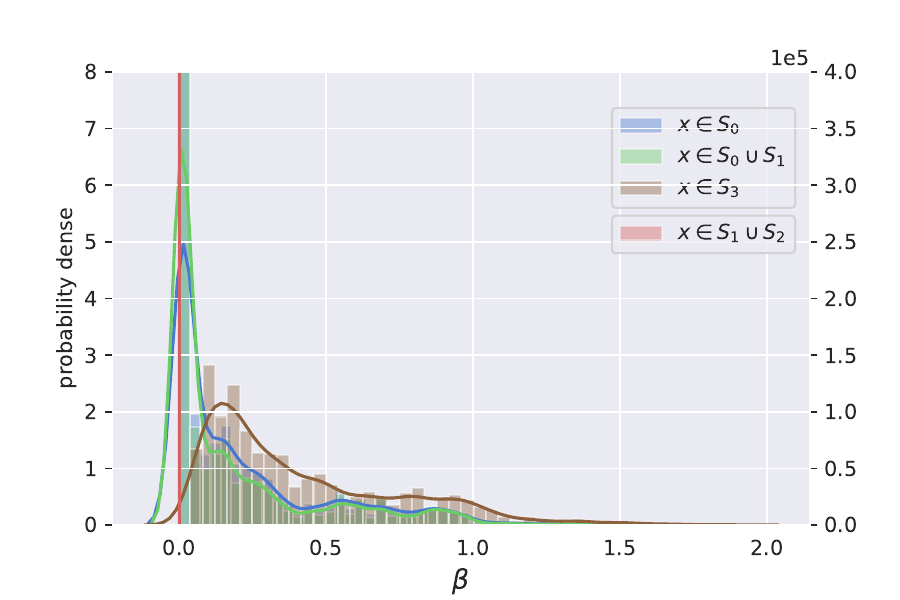}
        \label{fig:4c_beta}}
    \hfill
\subfloat[]{
        \includegraphics[width=0.45\linewidth,trim={0.2cm .1cm .7cm .8cm},clip]{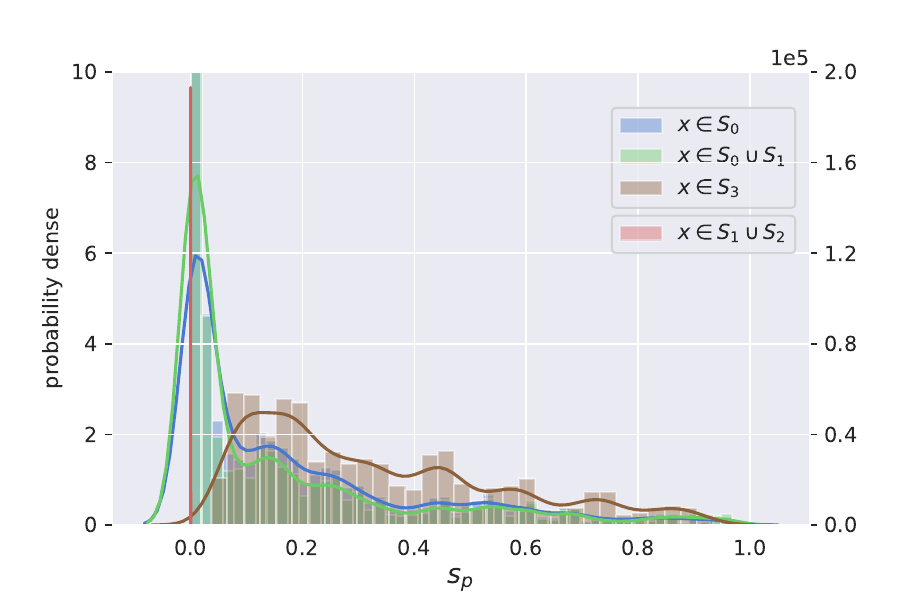}
        \label{fig:4c_psim}}

    \caption{Illustration of how $\text{Pr}(\beta>s_p)$ in Theorem 1 behaves in various cases by drawing negative pairs from different subsets of a 3-class feature space. The subsets are displayed in \Cref{fig:4c_sets}.  y-axis in the right side is for the case of $x \in S_1 \cup S_2$. The predictions are computed by the linear model. (a) Estimated distributions of $\beta$ when drawing negative pairs from different subsets. (b) Estimated distributions of $s_p$ when drawing negative pairs from different subsets.}
\label{fig:4c_demo}
\end{figure*}

We show the empirical distributions of $\beta$ and $s_p$ upon the four cases in \Cref{fig:4c_beta,fig:4c_psim}, respectively. In general, $s_p$ shows similar behaviors with $\beta$ in the four cases but in a smaller range ($\beta \in [0,2], s_p \in [0,1] $), which makes $\beta  > s_p$ tends to be true except when $\beta$ is around zero. Basically, a subset including more samples close to decision boundaries leads to more probability mass on larger values of $\beta$. In particular, the case of $S_3$ results in the largest probability of getting negative $\langle \bm{g_n}, \bm{g_m} \rangle$ because the predicted probabilities mostly concentrate on the two classes in a pair. 
These results explain that samples with most diverse gradients are close to decision boundaries because they tend to have high $\text{Pr}(\beta  > s_p)$ and $\langle \mathbf{x}_n, \mathbf{x}_m \rangle$ tend to be largely positive. 
Note that the gradients from $S_1 \cup S_2$ are not considered as diverse gradients despite that they are mostly negative. This is because $\langle \bm{\epsilon}_n, \bm{\epsilon}_m \rangle$ is close to zero in this case and thus $\langle \bm{g_n}, \bm{g_m} \rangle$ is close to zero too according to Lemma 1. And we can see that $\beta$ and $s_p$ both are around zero and $\text{Pr}(\beta  > s_p) = 0$ in this case. These results indicate that: \emph{smaller $\beta$ and $s_p$ lead to smaller $\text{Pr}(\beta  > s_p)$}.

\begin{figure*}[t!]
        \centering
\subfloat[]{
        \includegraphics[width=0.425\linewidth,trim={0.cm 0.cm 1.6cm 1.cm},clip]{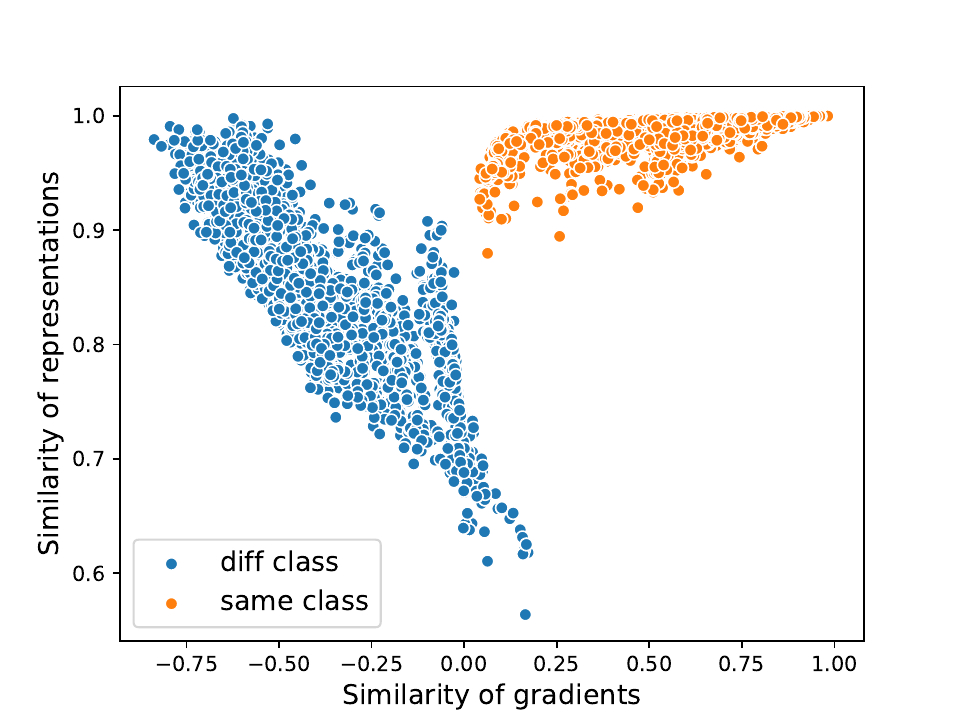}
        \label{fig:grads_corr_r_d79}}
        \hfill
\subfloat[]{
        \includegraphics[width=0.425\linewidth,trim={0.cm 0.cm 1.6cm 1.cm},clip]{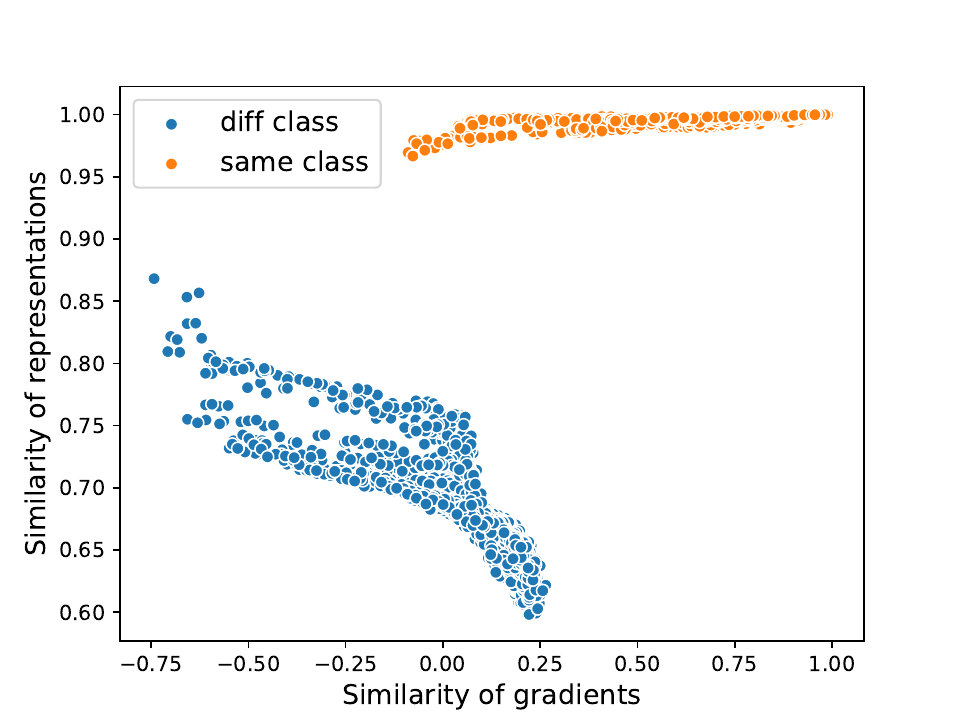}
        \label{fig:grads_corr_r_d01}}
        \caption{Similarities of gradients and representations of two classes in the MNIST dataset. The $x$ and $y$ axis are the cosine similarity of gradients and representations, respectively. Blue dots indicate the similarity of \emph{negative pairs} (two samples from different class), while orange dots indicate that of \emph{positive pairs} (two samples from a same class). (a) Class 7 \& 9. (b) Class 0 \& 1.}
        \label{fig:grads_corr_mnist}
    \vspace{-0.5cm}
    \end{figure*}

    \begin{table}[!t]
        \centering
        \caption{Demonstration of performance degradation in continual learning by compact representations. 
        }
        \begin{tabular}{c|c|c|c|c|c|c|c}
        \toprule
            \multirow{2}{*}{}          & \multirow{2}{*}{\begin{tabular}[c]{@{}c@{}}L1\\ (t=1)\end{tabular}} & \multirow{2}{*}{\begin{tabular}[c]{@{}c@{}}\# Act. Dim.\\ (t=1)\end{tabular}} & \multicolumn{5}{c}{Avg. Accuracy (in \%)} \\ \cline{4-8} 
                                       &                                                                     &                                                                                 & t=1   & t=2   & t=3  & t=4  & t=5  \\ \hline
            \multirow{2}{*}{MNIST}   & no                                                                  & 51                                                                              & 99.9  & 97.4  & 93.7 & 90.6 & 85.4 \\
                                       & yes                                                                 & 5                                                                               & 99.7  & 93.0  & 87.1 & 68.4 & 53.5 \\ \hline
            \multirow{2}{*}{Fashion} & no                                                                  & 66                                                                              & 98.3  & 90.3  & 83.1 & 74.4 & 77.3 \\
                                       & yes                                                                 & 8                                                                               & 97.4  & 88.9  & 75.7 & 56.1 & 50.2 \\ \bottomrule
            \end{tabular}
        \label{tab:l1_demo}
\end{table}

\section{The Relation between Gradients and Representations}

According to above findings, we consider two ways for reducing the gradient diversity: 1).minimizing inner products of  representations ($\langle \mathbf{x}_n, \mathbf{x}_m \rangle$) for negative pairs, which leads to less negative inner products of gradients; 2).minimizing inner products of predictions ($s_p$) for negative pairs, which decreases  $\text{Pr}(\beta  > s_p)$. Specifically, we concatenate the output logits with intermediate representations of a model and minimize inner products of them. 

We first verify this idea by training two binary classifiers for two groups of MNIST classes ($\{0,1\}$ and $\{7,9\}$). The classifiers have two hidden layers each with 100 hidden units and ReLU activations. We randomly chose 100 test samples from each group to compute the pairwise cosine similarities of representations and gradients. We display these similarities in \Cref{fig:grads_corr_r_d79,fig:grads_corr_r_d01}. The representations are obtained by concatenating outputs of all layers (including logits). The gradients are computed by all parameters of the model. In both figures we see that similarities of representations are all positive due to the ReLU activation. The similarities of gradients for positive pairs (orange dots) are mostly positive as well which is consistent with \Cref{thm2}. For negative pairs (blue dots), the classifier of class 0 and 1 gets smaller similarities of representations and much less negative similarities of gradients (\Cref{fig:grads_corr_r_d01}). It also gains a higher accuracy than the other classifier (99.95\% vs. 96.25\%), In all cases, the similarities of representations show strong correlations with the similarities of gradients (-0.86 and -0.85 for negative pairs, 0.71 and 0.79 for positive pairs).   These results illustrate the potential of reducing the gradient diversity by learning more discriminative representations between classes. 

\section{The Other Side of Discriminative Representations}

Learning more discriminative representations can be interpreted as learning larger margins between classes which has been an active research area for a few  decades. For example, \ac{KFD} \citep{mika1999fisher} and distance metric learning \citep{weinberger2006distance} aim to learn kernels that can obtain larger margins in an implicit representation space, whereas \acf{DML} \citep{kaya2019deep,roth2020revisiting} leverages deep neural networks to learn embeddings that maximize margins in an explicit representation space. 
Many popular methods in \ac{DML} aim to learn discriminative representations and are based on pairwise similarities (or distances) \citep{schroff2015facenet,wang2017deep,deng2019arcface,wang2019multi}. In principle they try to achieve this goal by minimizing the similarity (maximizing the distance) between different classes and maximizing the similarity (minimizing the distance) within a same class. 

According to previous sections, we see that minimizing the similarity  between different classes helps with reducing the  diversity of gradients. In this sense, approaches of \ac{DML} has the potential to help with \cl{}. 
However, the objective of these methods often includes the other side of learning discriminative representations:  learning compact representations within classes \citep{schroff2015facenet,wang2017deep,deng2019arcface}. The compactness results in sparser representations and smaller variance within classes, which means that information not useful for the classification task are likely to be omitted. Nonetheless, the unused information for the current task might be important for a future task in continual learning. For example, in the experiment of \Cref{fig:2dim_example} the y-dimension is not useful for the classification task but if there is a subsequent task adding a third class, as the green dots in \Cref{fig:4c_sets}, then y-dimension is necessary for  classification between the three classes. In such a case, even if we store previous samples into the episodic memory for future use, it  still introduces extra difficulties to generalize over all past tasks as the omitted dimensions can only be relearned by very limited samples in the memory. It could be much more difficult for high-dimensional data in particular. 

We verify this issue by training a model with and without L1 regulariztion at the first task of split-MNIST and split-Fashion MNIST (please refer to \Cref{sec:benchmark} for definitions of these tasks). In order to verify the influence of compact representations more preciely, we remove the L1 regularization in the later tasks so that the model is flexible to relearn ommitted dimensions without extra penalty. The episodic memory is formed by 300 samples that are uniformly randomly chosen from learned tasks. We identify active dimensions of the representation space after learning task 1 by selecting the hidden units that have a mean activation larger than 0.5 over all learned classes. The results are shown in \Cref{tab:l1_demo}, the average accuracy is computed over all learned classes. We see that with L1 regularization the model learns much more compact representations with a similar performance in the first task. However, the performance gets larger and larger degradation when more tasks have been encountered. 

The results demonstrate that  compact representations  may be detrimental to continual learning tasks. We need to restrain the compactness while learning discriminative representations, which is different with the common concept of \ac{DML}. \citet{roth2020revisiting} introduced a $\rho$-regularization method to prevent over-compression of representations.  The $\rho$-regularization method randomly replaces negative pairs by positive pairs in pairwise-based objectives with a pre-selected probability $p_{\rho}$. Nevertheless, these objectives still include the term for compactness. In addition, switching pairs is inefficient and may be detrimental to the performance in an online setting because some samples may never be learned in this way. 
Therefore, we propose an opposite way to \ac{DML} regarding the within-class compactness: \emph{\textbf{minimizing the similarities within classes for preserving information in the representation space}}.



%
%


\section{\acl{SDRL}}
Based on our findings in the above section, we now introduce an auxiliary objective \acf{SDRL} for classification tasks in \cl{}, which is straightforward and efficient.   
Instead of explicitly re-projecting gradients during training process, \ac{SDRL} helps with decreasing gradient diversity by optimizing the representations. As defined in \Cref{eq:drl}, \ac{SDRL} consists of two parts: one is for minimizing the inner products of representations from different classes ($\mathcal{L}_{bt}$) which forces the model to learn more discriminative representations and hence reduce the diversity of gradients; the other is for minimizing the inner products of representations from a same class ($\mathcal{L}_{wi}$) which helps preserve information for future tasks in \cl{}. We provide experimental results of an ablation study on \ac{SDRL} in \Cref{sec:ablat_drl}, according to which $\mathcal{L}_{bt}$ and $\mathcal{L}_{wi}$ both have shown effectiveness on improving the performance.
\begin{equation*}
     \begin{split}
         & \mathcal{L}_{SDRL} = \mathcal{L}_{bt} + \alpha \mathcal{L}_{wi}, \ \ \alpha > 0, \\
     \end{split}
 \end{equation*}
 \begin{equation}\label{eq:drl}
 \begin{split}
        & \mathcal{L}_{bt} = \frac{1}{N_{neg}}  \sum_{i=1}^B \sum_{j \neq i,y_j \neq y_i}^B \langle \bm{h}_{i},\bm{h}_{j}\rangle,\\
        &\mathcal{L}_{wi} = \frac{1}{N_{pos}} 
         \sum_{i=1}^B \sum_{j \neq i,y_j = y_i}^B \langle \bm{h}_{i},\bm{h}_{j}\rangle,
 \end{split}
 \end{equation}
where $B$ is training batch size. $N_{neg}, N_{pos}$ are the number of negative and positive pairs, respectively. $\alpha$ is a hyperparameter controlling the strength of $\mathcal{L}_{wi}$, $\bm{h}_{i}$ is the representation of $\mathbf{x}_i$, $\mathbf{y}_i$ is the label of $\mathbf{x}_i$. 
The final loss function combines the commonly used softmax  cross entropy loss for classification tasks ($\mathcal{L}$) with \ac{SDRL} ($\mathcal{L}_{SDRL}$) as shown in \Cref{eq:final_loss}, 
\begin{equation}\label{eq:final_loss}
    \begin{split}
        \widehat{\mathcal{L}} = \mathcal{L} + \lambda \mathcal{L}_{SDRL}, \ \ \lambda > 0,
    \end{split}
\end{equation}
where $\lambda$ is a hyperparameter controlling the strength of $ \mathcal{L}_{SDRL}$, which is larger for increased resistance to forgetting, and smaller for greater elasticity. 

The computational complexity of \ac{SDRL} is $O(B^2H)$, where $B$ is training batch size, $H$ is the dimension of representations. $B$ is usually small (not larger than 20) in related work with the online setting, and commonly $H \ll W$, where $W$ is the number of model parameters. 
 In comparison, the computational complexity of \ac{A-GEM} \citep{chaudhry2018efficient} and \ac{GSS}-greedy \citep{aljundi2019gradient} are $O(B_rW)$ and $O(BB_mW)$, respectively, where $B_r$ is the reference batch size in \ac{A-GEM} and $B_m$ is the memory batch size in \ac{GSS}. The computational complexity discussed here is additional to the cost of common back propagation. 
 We compare the training time of all methods in \Cref{tab:time} in \Cref{sec:experiments}, which shows the representation-based methods are much faster than gradient-based approaches. 

\section{Online memory update and \acl*{BER}}\label{sec:memory}

We follow the \emph{online setting} of \cl{} as was done for other gradient-based approaches with episodic memories \citep{lopez2017gradient,chaudhry2018efficient,aljundi2019gradient}, in which the model only trained with one epoch on the training data. 
We update the episodic memories by the basic ring buffer strategy: keep the last $n_c$ samples of class $c$ in the memory buffer, where $n_c$ is the memory size of a seen class $c$. 
We have deployed the episodic memories with a fixed size, implying a fixed budget for the memory cost. 
Further, we maintain a uniform distribution over all seen classes in the memory. The buffer may not be evenly allocated to each class before enough samples are acquired for newly arriving classes. Particularly, we directly load new data batches into the memory buffer without a separate buffer for the current task. The memory buffer works like a sliding window for each class in the data stream and we draw training batches from the memory buffer after loading new data batches from the data stream (i.e. new samples added into the memory buffer at each iteration). 
We show pseudo-code of the memory update strategy in \Cref{alg:ring_buf} for a clearer explanation. 

Since \ac{SDRL} and methods of \ac{DML} depend on the pairwise similarities of samples, we would prefer the training batch to include as wide a variety of different classes as possible to obtain sufficient discriminative information. Hence, we suggest a sophisticated sampling strategy of memory replay for the needs of such methods. The basic idea is to uniformly sample from all seen classes  to form a training batch, so that this batch can contain as many classes as possible. In addition, we ensure the training batch includes at least one positive pair to enable the parts computed by positive pairs in the loss. When the number of learned classes is much larger than the training batch size, this guarantee is necessary because the training batch may not include any positive pairs by just evenly sampling from all classes. Moreover, we also ensure the training batch includes classes in the current data batch so that it must include samples of the current task. We call this \acf{BER}. The pseudo code is in \Cref{alg:BER}. Note that we update  the memory and form the training batch based on the task ID instead of class ID for permuted MNIST tasks (please refer to \Cref{sec:benchmark} for definition of these tasks), as in this case each task always includes the same set of classes.   


  \begin{algorithm}[t]
    \caption{Ring Buffer Update with Fixed Buffer Size}
    \label{alg:ring_buf}
 \begin{algorithmic}
    \STATE {\bfseries Input:}
    \ \ $\mathbb{B}_t$ - current data batch, \\ \qquad \qquad $\mathbb{C}_t$ - the set of classes in $\mathbb{B}_t$,  
    \\ \qquad \qquad $\mathcal{M}$ - memory buffer, \\ \qquad \qquad $\mathbb{C}$ -  the set of classes in $\mathcal{M}$, \\ \qquad \qquad $K$ - memory buffer size. 
    \FOR{$c$ {\bfseries in} $\mathbb{C}_t$}
    \STATE Get $\mathbb{B}_{t,c}$ - samples of class $c$ in $\mathbb{B}_t$,\\ $\mathcal{M}_c$ - samples of class $c$ in $\mathcal{M}$,
    \IF{$c$ {\bfseries in} $\mathbb{C}$}
    {\STATE $\mathcal{M}_c = \mathcal{M}_c \cup \mathbb{B}_c$}
    \ELSE
    {\STATE $\mathcal{M}_c = \mathbb{B}_c$, \ \ 
    $\mathbb{C} = \mathbb{C} \cup \{c\}$}
    \ENDIF
    \ENDFOR
    \STATE
    $R =  |\mathcal{M}|+|\mathbb{B}| - K$
    \WHILE{$R > 0$}
    \STATE 
    $c{'} = \arg\max_{c}|\mathcal{M}_{c}|$
    \STATE remove the first sample in $\mathcal{M}_{c'}$,
    \STATE $R = R - 1$
    \ENDWHILE
    \STATE return $\mathcal{M}$ 
  \end{algorithmic}
  \end{algorithm}

\begin{algorithm}[t]
    \caption{\acl{BER}}
    \label{alg:BER}
 \begin{algorithmic}
    \STATE {\bfseries Input:}
    \ \ $\mathcal{M}$ - memory buffer, \\ \qquad \qquad $\mathbb{C}$ -  the set of classes in $\mathcal{M}$, \\ \qquad \qquad $B$ - training batch size, \\ \qquad \qquad $\bm{\theta}$ - model parameters, \\ \qquad \qquad $\mathcal{L}_{\bm{\theta}}$ - loss function, \\ \qquad \qquad $\mathbb{B}_t$ - current data batch, \\ \qquad \qquad $\mathbb{C}_t$ - the set of classes in $\mathbb{B}_t$, \\ \qquad \qquad $K$ - memory buffer size.  
 
    \STATE $\mathcal{M} \leftarrow $ MemoryUpdate($\mathbb{B}_t, \mathcal{C}_t, \mathcal{M}, \mathbb{C}, K$) 
    \STATE  $n_c, \mathbb{C}_{s},\mathbb{C}_{r} \leftarrow $ ClassSelection($\mathbb{C}_t, \mathbb{C}, B$)
    \STATE $\mathbb{B}_{train} = \emptyset$ 
    \FOR{$c$ {\bfseries in} $\mathbb{C}_s$}
    \IF {$c$ {\bfseries in} $\mathbb{C}_r$}
    \STATE $m_c = n_c + 1$
    \ELSE
    \STATE $m_c = n_c$
    \ENDIF
    \STATE Get $\mathcal{M}_c$ $\lhd$ samples of class $c$ in $\mathcal{M}$,
    \STATE $\mathbb{B}_c \overset{m_c}{\sim} \mathcal{M}_c$ $\lhd$ sample $m_c$ samples from $\mathcal{M}_c$
    \STATE $\mathbb{B}_{train} = \mathbb{B}_{train} \cup \mathbb{B}_c$
    \ENDFOR
    \STATE $\bm{\theta} \leftarrow \text{Optimizer}(\mathbb{B}_{train},\bm{\theta}, \mathcal{L}_{\bm{\theta}})$
    \\ \quad 
  \end{algorithmic}
  \end{algorithm}

\begin{algorithm}[tbh!]
    \caption{Class Selection for \ac{BER}}
    \label{alg:class_select}
 \begin{algorithmic}
    \STATE {\bfseries Input:}
    \ \ $\mathbb{C}_t$ - the set of classes in current data batch $\mathbb{B}_t$, \\ \qquad \qquad $\mathbb{C}$ -  the set of classes in the memory $\mathcal{M}$,  \\ \qquad \qquad $B$ - training batch size, \\ \qquad \qquad $m_{p}$ - minimum number of positive pairs  ($m_{p} \in \{0,1\}$) .  
    \STATE 
    $n_c = \lfloor B/|\mathbb{C}|\rfloor,\ \ r_c =  B \mod {|\mathbb{C}|} $, 
    \IF{$B > |\mathbb{C}|$ or $m_p == 0$}
    \STATE $\mathbb{C}_{r} \overset{r_c}{\sim} \mathbb{C}$  \ \ \ $\lhd$ sample $r_c$ classes from all seen classes without replacement.
    \STATE $\mathbb{C}_s = \mathbb{C}$
    \ELSE 
    \STATE $\mathbb{C}_{r} = \emptyset$, $n_c = 1$, $n_s = B - |\mathbb{C}_t|$, $\lhd$ ensure the training batch including samples from the current task. 
    \STATE $\mathbb{C}_s \overset{n_s-1}{\sim} (\mathbb{C}-\mathbb{C}_t)$ $\lhd$ sample $n_s - 1$ classes from all seen classes except classes in $\mathbb{C}_t$.
    \STATE $\mathbb{C}_s = \mathbb{C}_s \bigcup \mathbb{C}_t$, 
    \STATE $\mathbb{C}_{r} \overset{1}{\sim} \mathbb{C}_s$
    $\lhd$ sample one class to have a positive pair

    \ENDIF
    \STATE {\bfseries Return:} $n_c, \mathbb{C}_{s},\mathbb{C}_{r}$
  \end{algorithmic}
  \end{algorithm}

\section{Experimental Results}\label{sec:experiments}

In this section we evaluate our methods on multiple benchmark tasks by comparing with several baseline methods in the setting of online continual learning. We also provide comprehensive results of an ablation study on \ac{SDRL} which obtain more insights of our method. The source code of our experiments are available at \url{https://github.com/yc14600/discriminative-representation-loss}.

\subsection{Benchmark tasks}\label{sec:benchmark}
The benchmark tasks we applied in our experiments are listed below, where the tasks of MNIST and CIFAR-10 datasets follow the setting in \citet{aljundi2019gradient}: 

\emph{Permuted MNIST}: 10 tasks using the MNIST dataset \citep{lecun2010mnist}, each task includes the same 10 classes with different permutation of features (784 pixels in this case). The training size is 1000 samples per task, memory size is 300 samples;
    
 \emph{Split MNIST}: 5 tasks using the MNIST dataset, each task includes two classes which are disjoint from the other tasks. The training size is 1000 samples per task, memory size is 300 samples;

 \emph{Split Fashion-MNIST}: 5 tasks using the Fashion-MNIST dataset \citep{xiao2017/online},the same setting as \emph{Split MNIST};
    
 \emph{Split CIFAR-10}: 5 tasks using the CIFAR-10 dataset \citep{krizhevsky2009learning},  each task includes two classes which are disjoint from other tasks. The training size is 2000 samples per task, memory size is 1000 samples;
    
 \emph{Split CIFAR-100}: 10 tasks using the CIFAR-100 dataset \citep{krizhevsky2009learning},  each task includes 10 classes which are disjoint from other tasks. The training size is 5000 samples per task, memory size is 5000 samples.

 \emph{Split TinyImageNet}: 20 tasks using the TinyImageNet dataset \citep{le2015tiny},  each task includes 10 classes which are disjoint from other tasks. The training size is 5000 samples per task, memory size is 5000 samples.

\subsection{Baselines} 
We compare our methods with: two gradient-based approaches (\emph{\ac{A-GEM}} \citep{chaudhry2018efficient} and \emph{\ac{GSS}-greedy} \citep{aljundi2019gradient}), two standalone experience replay methods (\emph{\ac{ER}} \citep{chaudhry2019continual} and \emph{\ac{BER}}), two \acs{SOTA} methods of \ac{DML} (\emph{Multisimilarity} \citep{wang2019multi} and  \emph{R-Margin} \citep{roth2020revisiting}). We have introduced \ac{A-GEM} and \ac{GSS} in \Cref{sec:related}. In the following we give a brief introduction of other baselines.
%

          

\emph{\ac{ER}} \citep{chaudhry2019continual}: a basic experience replay strategy, yet was shown to achieve better performance than \ac{A-GEM} and \ac{MER} \citep{riemer2018learning} in the online continual learning setting. It simply composes a training batch divided equally between samples from the episodic memory and samples from the current task. We consider this as a baseline of replay-based methods.

\emph{Multisimilarity} \citep{wang2019multi}: A method of \ac{DML} which has shown outstanding performance in a comprehensive empirical study of \ac{DML} \citep{roth2020revisiting}. We adopt the loss function of Multisimilarity as an auxiliary objective in classfication tasks of \cl{}, the batch mining process is omitted since we have access to true labels of the training set. So the loss function is $\widehat{\mathcal{L}} = \mathcal{L} + \lambda \mathcal{L}_{mul}$, and:
\begin{equation}\label{eq:mults}
    \begin{split}
        &\mathcal{L}_{mul} =  \frac{1}{B} \sum_{i=1}^B \left[\frac{1}{\alpha} \log [1+ \sum_{y_j=y_i} \exp{(-\alpha (s_c(\bm{h}_i,\bm{h}_j)-\gamma)})] \right. \\
        & \qquad \left. + \frac{1}{\beta}\log{[1+\sum_{y_j \neq y_i}\exp{(\beta (s_c(\bm{h}_i,\bm{h}_j)-\gamma))}}]\right]
    \end{split}
\end{equation}
where $s_c(\cdot,\cdot)$ is cosine similarity, $\alpha, \beta,\gamma$ are hyperparameters. In all of our experiments we set $\alpha=2, \beta=40, \gamma=0.5$ as the same as in \citet{roth2020revisiting}. 

\emph{R-Margin} \citep{roth2020revisiting}: A  method of \ac{DML} which deploy the $\rho$ regularization method for Margin loss \citep{wu2017sampling} and has shown outstanding performance in \citep{roth2020revisiting} as well. We similarly deploy R-Margin for \cl{} as an auxiliary objective too, which uses the Margin loss \citep{wu2017sampling} with the $\rho$ regularization \citep{roth2020revisiting}. So the loss function is $\widehat{\mathcal{L}} = \mathcal{L} + \lambda \mathcal{L}_{margin}$, and:
\begin{equation}\label{eq:rmargin}
    \begin{split}
        \mathcal{L}_{margin} &= \sum_{i=1}^B \sum_{j=1}^B \gamma + \mathbb{I}_{j\neq i, y_j=y_i} (d(\bm{h}_i,\bm{h}_j)-\beta) \\
        & \qquad - \mathbb{I}_{y_j \neq y_i}(d(\bm{h}_i,\bm{h}_j)-\beta)
    \end{split}   
\end{equation}
where $d(\cdot,\cdot)$ is Euclidean distance, $\beta$ is a trainable variable and $\gamma$ is a hyperparameter. We follow the setting in \citet{roth2020revisiting}: $\gamma=0.2$, the initialization of $\beta$ is $0.6$, and $p_{\rho} = 0.2$ in $\rho$ regularization. 

 \emph{\textbf{N.B.}}: We deploy the losses of Multisimilarity and R-Margin as auxiliary objectives as the same as \ac{SDRL} because using standalone such losses causes difficulties of convergence in the settings of online \cl{}.
 
\subsection{Performance measures}\label{sec:measure}

We use the following measures to evaluate the performance of all methods:

\emph{Average accuracy}, which is evaluated after learning all tasks: 
\begin{equation}
    \bar{a}_t = \frac{1}{t} \sum_{i=1}^t a_{t,i}
\end{equation}
where $t$ is the index of the latest task, $a_{t,i}$ is the accuracy of task $i$ after learning task $t$.
    
\emph{Average forgetting} \citep{chaudhry2018riemannian}, which measures average accuracy drop of all tasks after learning the whole task sequence: 
\begin{equation}
    \bar{f}_t = \frac{1}{t-1} \sum_{i=1}^{t-1} \max_{j \in \{i,\dots,t-1\}} (a_{j,i} - a_{t,i})
\end{equation}

\emph{Average intransigence} \citep{chaudhry2018riemannian}, which measures the inability of a model learning new tasks: 
\begin{equation}
  \bar{I}_t = \frac{1}{t}\sum_{i=1}^t a_i^* - a_{i,i},  
\end{equation}
where $a_i^*$ is the upper-bound accuracy of task $i$ that could be obtained by a model trained solely on this task. Instead of training an extra model for each task, we use the best accuracy among all compared models as $a^*_i$ which does not require additional computational overhead.  

\setlength{\tabcolsep}{3pt}
\begin{table*}[tbh!]
\centering
    \caption{Average accuracy (in \%) over all tasks after learning the whole sequence, the higher the better.}
\begin{tabular}{ccccccc}
    \toprule
                & P-MNIST             & S-MNIST                & Fashion              &  CIFAR10              &  CIFAR100 
                &
                TinyImageNet
                \\
\midrule
SDRL             & $\mathbf{80.5 \pm 0.4}$ & $\mathbf{88.1 \pm 0.6}$ & $\mathbf{77.9 \pm 0.8}$ & $\mathbf{40.4 \pm 1.5}$          & $\mathbf{19.3 \pm 0.5}$ & $\mathbf{8.3 \pm 0.2}$ \\
BER             & $79.2 \pm 0.3$          & $85.2 \pm 1.1$          & $77.0 \pm 0.7$          & $37.3 \pm 1.4$          & $18.2 \pm 0.4$   & $6.7 \pm 0.8$       \\
ER              & $78.2 \pm 0.6$          & $83.2 \pm 1.5$           & $75.8 \pm 1.4$          & $39.4 \pm 1.6$           & $18.3 \pm 0.3$   & $7.6 \pm 0.6$       \\
A-GEM           & $76.7 \pm 0.5$          & $84.5 \pm 1.1$          & $66.4 \pm 1.5$          & $25.7 \pm 3.3$          & $16.5 \pm 1.2$    & $2.0 \pm 0.8$      \\
GSS      & $77.1 \pm 0.3$          & $82.8 \pm 1.8$          & $72.5 \pm 0.9$          & $33.6 \pm 1.7$            & $13.9 \pm 1.0$     & $3.3 \pm 0.2$      \\
Multisim & $79.5 \pm 0.6$          & ${86.3 \pm 1.1}$ & $77.2 \pm 0.6$          & ${38.9 \pm 3.2}$ & $18.4 \pm 0.5$    & $6.8 \pm 1.0$      \\
R-Margin        & $78.0 \pm 0.3$          & $85.6 \pm 0.9$           & $76.6 \pm 0.8$           & $36.7 \pm 1.7$          & $18.3 \pm 0.5$   & $6.4 \pm 0.3$   \\ 
\bottomrule    
\end{tabular}
    \label{tab:acc}
\end{table*}

\begin{table*}[tbh!]
\centering 
    \caption{Average forgetting (in \%), measuring average accuracy drop at the end of the sequence, the lower the better.}
    \begin{tabular}{ccccccc}
        \toprule
                        & P-MNIST             & S-MNIST                & Fashion              & CIFAR10              & CIFAR100
                        &
                        TinyImageNet
                        \\
        \midrule
        SDRL             & $\mathbf{4.5 \pm 0.2}$          & $\mathbf{8.9 \pm 0.7}$ & $\mathbf{16.6 \pm 1.9}$          & $41.4 \pm 3.2$          & ${29.0 \pm 1.0}$ & $27.0 \pm 1.4$ \\
        BER             & $5.1 \pm 0.3$          & $13.0 \pm 1.5$          & $18.2 \pm 2.7$          & $52.3 \pm 1.6$          & $35.9 \pm 1.0$  & ${26.0 \pm 1.2}$        \\
        ER              & $7.1 \pm 0.6$          & $17.2 \pm 1.9$          & $24.0 \pm 2.7$          & $50.3 \pm 1.9$          & $37.0 \pm 1.5$ & $28.5 \pm 2.0$         \\
        A-GEM           & ${5.4 \pm 0.4}$ & $12.6 \pm 1.3$          & $37.0 \pm 1.9$          & $40.4 \pm 4.6$          & ${25.3 \pm 1.4}$    & $24.4 \pm 0.6$      \\
        GSS      & $7.6 \pm 0.2$          & $ 17.9\pm 2.4$          & $27.4 \pm 2.2$          & $\mathbf{27.6 \pm 4.0}$ & $\mathbf{18.6 \pm 0.7}$     & $\mathbf{11.3 \pm 0.7}$     \\
        Multisim & $5.0 \pm 0.7$          & $12.0 \pm 1.4$          & $18.8 \pm 2.2$          & $48.0 \pm 3.8$          & $35.8 \pm 0.6$    &  $37.6 \pm 0.4$      \\
        R-Margin        & $5.4 \pm 0.3$          & $12.5 \pm 1.4$          & ${17.1 \pm 3.0}$ & $50.5 \pm 2.6$          & $35.1 \pm 0.5$  & $38.1 \pm 0.7$\\ \bottomrule      
        \end{tabular}
        \label{tab:fgt}
\end{table*}

\begin{table*}[tbh!]
\centering
\caption{Average intransigence (in \%), measuring the inability of learning a new task, the lower the better.}
\begin{tabular}{ccccccc}
    \toprule
             & P-MNIST             & S-MNIST                 & Fashion              & CIFAR10              & CIFAR100 & TinyImageNet            \\
    \midrule
    SDRL      & ${2.0 \pm 0.5}$ & $2.8 \pm 0.3$           & $6.9 \pm 1.0$          & $9.9 \pm 1.5$          & $9.0 \pm 1.0$       & $25.0 \pm 7.5$   \\
    BER      & $3.0 \pm 0.2$          & $2.5 \pm 0.3$           & $7.7 \pm 1.2$          & $4.3 \pm 0.6$          & $4.0 \pm 0.6$   & $28.3 \pm 0.5$       \\
    ER       & $\mathbf{1.8 \pm 0.4}$          & $1.2 \pm 0.1$           & $4.1 \pm 0.6$          & $\mathbf{3.8 \pm 1.0}$ & $\mathbf{2.8 \pm 1.0}$ & ${11.8 \pm 2.5}$\\
    A-GEM    & $7.0 \pm 0.7$          & $3.5 \pm 0.3$           & $\mathbf{1.0 \pm 0.3}$ & $25.4 \pm 2.1$          & $15.2 \pm 1.0$     & $21.4 \pm 1.3$     \\
    GSS      & $7.6 \pm 0.2$          & $\mathbf{0.8\pm 0.3}$ & $27.4 \pm 2.2$          & $27.9 \pm 2.0$          & $22.3 \pm 1.4$   & $45.0 \pm 0.1$       \\
    Multisim & $2.6 \pm 0.2$          & $2.2 \pm 0.3$           & $6.2\pm 1.0$           & $6.1 \pm 1.1$          & $3.8 \pm 0.6$     & $4.0 \pm 1.2$     \\
    R-Margin & $3.6 \pm 0.4$          & $2.4 \pm 0.3$           & $10.5 \pm 1.7$          & $6.3 \pm 2.1$          & $4.5 \pm 0.7$  & $\mathbf{3.9 \pm 0.4}$  \\ \bottomrule     
    \end{tabular}
            \label{tab:itrs}
\end{table*}

\subsection{Experimental settings}
Besides online training, we use \textbf{\emph{single-head}} (shared output) models in all of our experiments, meaning that we do not require the task identifier at testing time. This setting is more practical in real applications but  more difficult for continual learning.
We use the vanilla SGD optimizer for all experiments without any scheduling. For tasks on MNIST and Fashion-MNIST, we use a \ac{MLP} with two hidden layers (100 neurons per layer) and ReLU activations. For tasks on CIFAR datasets and TinyImageNet, we use the same reduced Resnet18 as used in \citet{chaudhry2018efficient,aljundi2019gradient}. All networks are trained from scratch without any preprocessing or augmentation of data except standardization. For all models, representations are the concatenation of outputs of all dense layers (which including logits). 
We deploy \ac{BER} as the replay strategy for \ac{SDRL}, Multisimilarity, and R-Margin. 
 
To make a fair comparison of all methods, the configurations of GSS-greedy are as suggested in \citep{aljundi2019gradient}, with batch size set to 10 and each batch receives multiple iterations. 
For the other methods, we use the ring buffer memory as described in \Cref{alg:ring_buf}, the loading  batch size is set to 1, following with one iteration. As we follow the online setting of training on limited  data with one epoch, we either use a small batch size or iterate on one batch several times to obtain necessary steps for gradient optimization. We chose a small batch size with one iteration instead of larger batch size with multiple iterations because by our memory update strategy (\Cref{alg:ring_buf}) it achieves similar performance without tuning the number of iterations. Since \ac{GSS}-greedy has a different strategy for updating memories, we use its reported settings in \citep{aljundi2019gradient}. 

Details of other hyperparameters are given in \Cref{tab:hyper}.
We use 10\% of training set as validation set for choosing hyperparameters by cross validation. The standard deviation shown in all results are evaluated over 10 runs with different random seeds.

\subsection{Comparison with  baselines}
\Cref{tab:acc,tab:fgt,tab:itrs} give the average accuracy, forgetting, and intransigence of all methods on all benchmark tasks, respectively. 
TinyImageNet gets much worse performance than other benchmarks because it is the most difficult one as it has more classes (200), a longer task sequence (20 tasks), and higher feature dimensions ($64 \times 64 \times 3$). In comparison, CIFAR datasets have $32 \times 32 \times 3$ features,  MNIST and Fashion-MNIST have $28 \times 28 \times 1$ features.

As we can see, the forgetting and intransigence often contradictory to each other which is a common phenomenon in \cl{} due to the trade-off between the plasticity and stability of the model. \ac{SDRL} is able to get a better trade-off between them and thus outperforms other methods over most benchmark tasks in terms of average accuracy. We provide an ablation study of the two components in \ac{SDRL} in \Cref{tab:ablat_drl}  which shows \ac{SDRL} obtains an improved intransigence by $\mathcal{L}_{wi}$ and an improved forgetting by $\mathcal{L}_{bt}$ in most cases, the two terms are complementary to each other and combining them brings benefits on both sides.
Multisimilarity and R-Margin both have shown relatively good performance, which shows learning more discriminative representations can be a more efficient way than directly refining gradients.

\Cref{tab:cifar_time} compares the training time of all methods on several benchmarks. We can see that simply memory-replay methods \acs{ER} and \acs{BER} are faster than others; representation-based methods \acs{SDRL}, Multisimilarity, and R-Marigin take similar training time with memory-replay methods; the gradient-based methods \acs{A-GEM} and \acs{GSS} are much slower than others, especially on a larger model. The experiments with the \acs{MLP} have been tested on a laptop with an 8-core Intel CPU and 32G RAM, the experiments with the reduced Resnet18 have been tested on a server with a NVIDIA TITAN V GPU.
\setlength{\tabcolsep}{3pt}
\begin{table*}[tbh!]
    \centering
    \small
    \caption{Training time (in seconds) of the whole task sequence of several benchmarks.}
    \label{tab:time}
    \begin{tabular}{cccccccc}
    \toprule
                   & \multicolumn{1}{c}{ SDRL } & \multicolumn{1}{c}{ BER }
                   &             \multicolumn{1}{c}{ ER }   
                   &       \multicolumn{1}{c}{ A-GEM }
                   &       \multicolumn{1}{c}{GSS}
                   &       \multicolumn{1}{c}{Multisim}
                   &       \multicolumn{1}{c}{R-Margin}
                   \\
    \midrule
    P-MNIST (MLP) & ${12.48 \pm 0.16}$           & $11.17 \pm 0.18$
    & $\mathbf{10.38 \pm 0.05}$
    & $28.0 \pm 0.09$
    & $33.98 \pm 0.6$
    & $12.91 \pm 0.13$
    & $13.45 \pm 0.14$
    \\   
    S-MNIST (MLP)   & ${5.6 \pm 0.14}$      
    & $5.29 \pm 0.08$ 
    & $\mathbf{5.25 \pm 0.02}$ 
    &  $13.41 \pm 0.47$ 
    &  $19.07 \pm 1.31$ 
    & $5.89 \pm 0.09$
    & $6.29 \pm 0.43$
    \\   
    CIFAR10 (r. Resnet18)  & ${265.1 \pm 0.9}$      
    & $264.3 \pm 0.6$ 
    & $\mathbf{261.5 \pm 0.3}$ 
    &  $1067.1 \pm 5.6$ 
    &  $5289.4 \pm 7.3$ 
    & $286.4 \pm 0.7$
    & $281.7 \pm 0.8$
    \\
    \bottomrule
    \end{tabular}
    \label{tab:cifar_time}
    \end{table*} 

We train models from scratch in our experiments to verify the sheer effectiveness of our methods. In many computer vision applications, a pre-trained model is often deployed with fine-tuning for a specific task. \ac{SDRL} can perform well in such scenarios as well. \ac{SDRL} is a wining solution for the CLVision Challenge \footnote{hold by the Continual Learning workshop in CVPR 2020,  \url{https://sites.google.com/view/clvision2020/challenge?authuser=0}} and shows consistently good performance on all benchmarks \citep{lomonaco2021cvpr}. In this challenge we used a ResNeSt50 \citep{zhang2020resnest} pre-trained on ImageNet \citep{ILSVRC15} without extra preprocessing or augmentation of data.

\subsection{Ablation study on \ac{SDRL}}
\label{sec:ablat_drl}

\begin{figure*}[htb]
        \centering
\subfloat[]{
        \includegraphics[width=0.31\linewidth,trim={0.6cm 0.cm 1.5cm 1.3cm},clip]{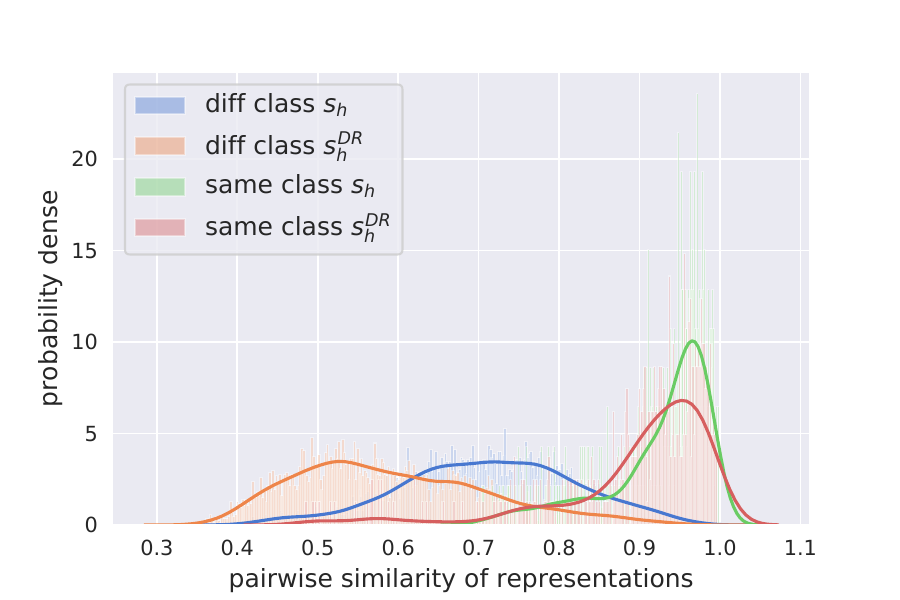}
        \label{fig:rep_sim}}
        \hfill%
\subfloat[]{
        \includegraphics[width=0.31\linewidth,trim={0.6cm 0.cm 1.5cm 1.3cm},clip]{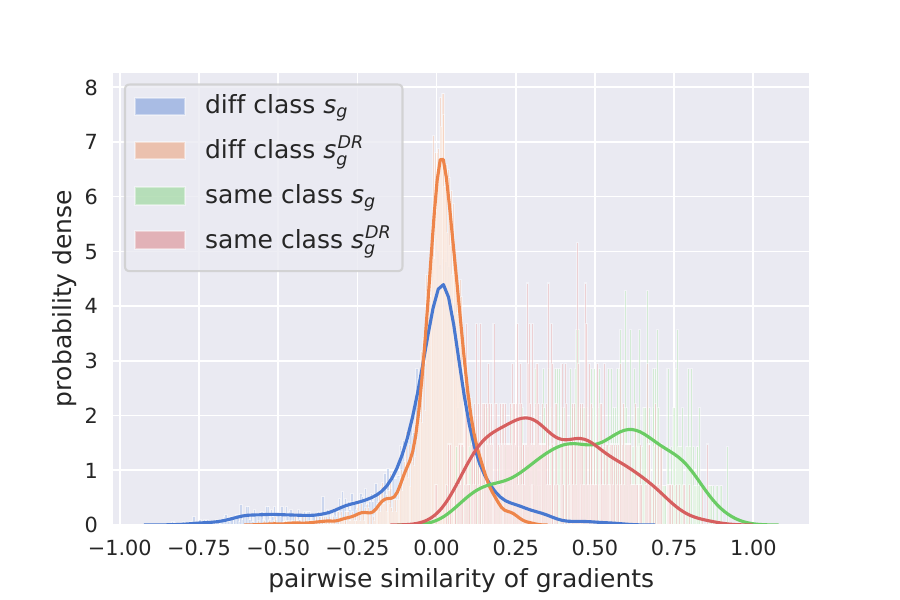}
        \label{fig:grad_sim}}
        \hfill%
\subfloat[]{
        \includegraphics[width=0.31\linewidth,trim={0.6cm 0.cm 1.5cm 1.3cm},clip]{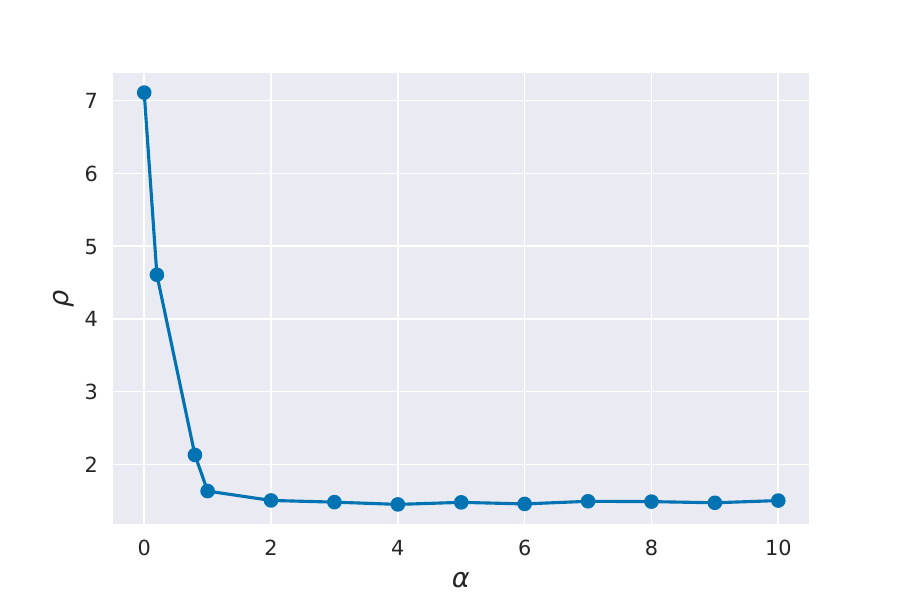}
        \label{fig:rho_relation}}
        \caption{Effects of $\mathcal{L}_{SDRL}$ on reducing diversity of gradients and $\rho$-spectrum. (a) and (b) display empirical distributions of similarities of representations and gradients, respectively. $s_h^{DR}$ and $s_h$ denote similarities of representations with and without $\mathcal{L}_{SDRL}$, respectively, $s_g^{DR}$ and $s_g$ denote similarities of gradients with and without $\mathcal{L}_{SDRL}$, respectively. (c) demonstrates increasing $\alpha$ in $\mathcal{L}_{SDRL}$ can decrease $\rho$-spectrum effectively.}
        \label{fig:obj_verify}
    \end{figure*}
    \begin{figure*}[h!]
            \centering
            \subfloat[]{          \includegraphics[width=0.42\linewidth,trim={0.2cm .1cm 1.6cm 1.cm},clip]{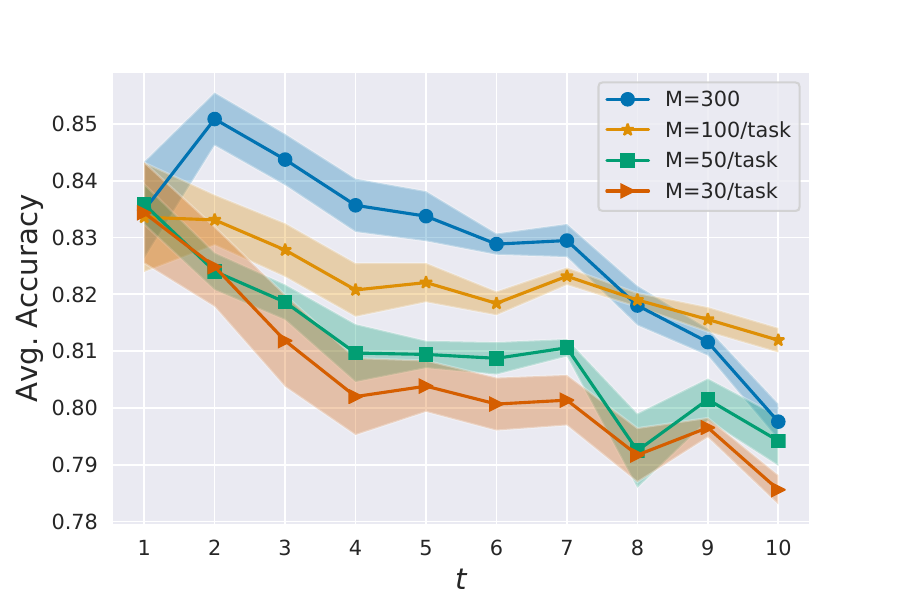}
            \label{fig:permuted_msize}}
            \hfill
\subfloat[]{            \includegraphics[width=0.42\linewidth,trim={0.1cm .1cm 1.6cm 1.cm},clip]{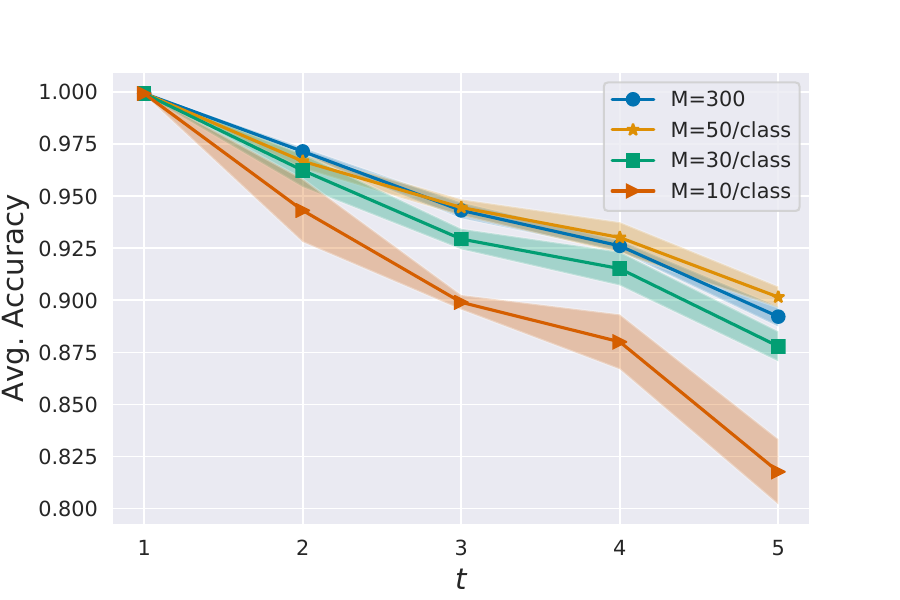}
            \label{fig:split_msize}}
            \caption{Average accuracy of \ac{SDRL} with different memory budgets. The $x$ axis is the index of tasks, the $y$ axis is the average accuracy over learned tasks. The shaded area is plotted by standard deviation of 10 runs. (a) Results of Permuted MNIST. (b) Results of Split MNIST.}
            \label{fig:msize}
        \end{figure*}

We have conducted a series of experiments as an ablation study to obtain more insights on \ac{SDRL}, including verifying the effects of \ac{SDRL} on reducing the diversity of gradients and preserving more information on representations, comparing the two terms of within/between-classes in \ac{SDRL}, comparing different memory-replay strategies and memory budgets with \ac{SDRL}.

\emph{Verifying the effects of \ac{SDRL}}: We verify the expected effects of $\mathcal{L}_{SDRL}$ by training a model with/without $\mathcal{L}_{SDRL}$ on  Split-MNIST tasks: \Cref{fig:rep_sim} shows that $\mathcal{L}_{SDRL}$ notably reduces the similarities of representations from different classes while making representations from a same class less similar; \Cref{fig:grad_sim} shows more of the gradient similarity of different classes are pushed to zero and hence less probability mass over negative ones. These results prove that \ac{SDRL} can achieve our goal in reducing diversity of gradients. The phenomenon of within-class similarities is caused by $\mathcal{L}_{wi}$ in \ac{SDRL}. To quantify the information preserved in representations, we apply \emph{$\rho$-spectrum} \citep{roth2020revisiting} to measure the information entropy contained in the representation space. It computes the KL-divergence between a discrete uniform distribution and the spectrum of data representations. Lower values of $\rho$-spectrum indicate higher variance of the representations and hence more information entropy retained.
\Cref{fig:rho_relation} demonstrates increasing $\alpha$ (the weight of $\mathcal{L}_{wi}$) can effectively decrease $\rho$-spectrum to a low-value level, which matches our expectation on $\mathcal{L}_{wi}$. However, $\rho$-spectrum is not always the smaller the better because it may retain too much noise when the information entropy is too high. In our experiments, we found that setting $\alpha$ to 1 or 2 is good enough for most cases.

\begin{table*}[t]
    \centering
    \caption{Comparing the performance with or without the two terms ($\mathcal{L}_{bt}$, $\mathcal{L}_{wi}$) in \ac{SDRL}. All criteria are in percentage. The bold font indicates the best performance of a criterion.}
    \begin{tabular}{c|c|c|c|c|c}
    \hline
                             &      & BER              & \multicolumn{3}{c}{SDRL}                                                     \\ \hline
                             &      &    with none              & with $\mathcal{L}_{bt}$ & with $\mathcal{L}_{wi}$ & with both                      \\ \hline
    \multirow{4}{*}{P-MNIST} & Avg. Accuracy  & $79.2 \pm 0.3$    & $79.6 \pm 0.4$        & $80.1 \pm 0.4$        & $\mathbf{80.5 \pm 0.4}$    \\
                             & Avg. Forgetting & $5.1 \pm 0.3$   & $4.8 \pm 0.3$         & $4.9 \pm 0.5$         & $\mathbf{4.5 \pm 0.2}$       \\
                             & Avg. Intransigence & $2.9 \pm 0.2$ & $2.6 \pm 0.2 $        & $\mathbf{1.9 \pm 0.3} $        & $\mathbf{1.9 \pm 0.5}$      \\
                             \hline
    \multirow{4}{*}{S-MNIST} & Avg. Accuracy   & $85.2 \pm 1.1$   & $86.5 \pm 0.9$        & $86.6 \pm 0.7$        & $\mathbf{88.1 \pm 0.6}$    \\
                             & Avg. Forgetting & $13.0 \pm 1.5$   & $11.3 \pm 1.3$        & $11.7 \pm 1.1$        & $\mathbf{8.9 \pm 0.7}$     \\
                             & Avg. Intransigence & $1.4 \pm 0.3$ & $1.4 \pm 0.3$         & $\mathbf{1.0 \pm 0.2}$         & $1.7 \pm 0.3$      \\
                             \hline
    \multirow{4}{*}{Fashion} & Avg. Accuracy  & $77.0 \pm 0.7$    & $77.3 \pm 0.8$        & $77.1 \pm 0.8$        & $\mathbf{77.9 \pm 0.8}$    \\
                             & Avg. Forgetting  & $18.2 \pm 2.7$   & $18.1 \pm 2.4$        & $18.5 \pm 2.2$        & $\mathbf{16.6 \pm 1.9}$    \\
                             & Avg. Intransigence & $3.7 \pm 1.8$ & $3.6 \pm 1.5$         & $\mathbf{3.4 \pm 1.1}$         & $4.1 \pm 1.4$      \\
                             \hline
    \multirow{4}{*}{CIFAR10} & Avg. Accuracy  & $37.3 \pm 1.4$    & $\mathbf{40.4 \pm 1.8}$        & $39.4 \pm 0.8$        & $\mathbf{40.4 \pm 1.6}$    \\
                             & Avg. Forgetting  & $52.3 \pm 1.6$   & $45.7 \pm 2.1$        & $46.6 \pm 2.1$        & $\mathbf{41.4 \pm 3.2}$    \\
                             & Avg. Intransigence & $\mathbf{3.5 \pm 0.6}$ & $5.6 \pm 1.0$         & $5.9 \pm 1.8$         & $9.0 \pm 1.5$      \\
                             \hline
     \multirow{4}{*}{CIFAR100} & Avg. Accuracy  & $18.2 \pm 0.4$    & ${18.7 \pm 0.3}$        & $18.5 \pm 0.4$        & $\mathbf{19.3 \pm 0.5}$    \\
                             & Avg. Forgetting  & $35.9 \pm 1.0$   & $34.7 \pm 0.5$        & $36.2 \pm 0.5$        & $\mathbf{29.0 \pm 1.0}$    \\
                             & Avg. Intransigence & ${3.1 \pm 0.6}$ & $3.7 \pm 0.3$         & $\mathbf{2.5 \pm 0.4}$         & $8.1 \pm 1.0$      \\
                             \hline
     \multirow{4}{*}{TinyImageNet} & Avg. Accuracy  & $6.7 \pm 0.8$    & ${7.2 \pm 0.5}$        & $6.7 \pm 0.4$        & $\mathbf{8.3 \pm 0.2}$    \\
                             & Avg. Forgetting  & $\mathbf{26.0 \pm 1.2}$   & $39.6 \pm 0.7$        & $40.4 \pm 0.3$        & ${27.0 \pm 1.4}$    \\
                             & Avg. Intransigence & ${17.3 \pm 1.9}$ & $3.9 \pm 1.0$         & $\mathbf{3.7 \pm 0.5}$         & $14.8 \pm 1.5$      \\
                             \hline
    \end{tabular}
    \label{tab:ablat_drl}
\end{table*}

\emph{Comparing the two terms of \ac{SDRL}}:  
\Cref{tab:ablat_drl} provides the results of comparing the effects of the two terms in \ac{SDRL}. In general, both of them show improvements on standalone \ac{BER} in most cases. $\mathcal{L}_{bt}$ gets more improvements on forgetting, $\mathcal{L}_{wi}$ gets more improvements on intransigence. Overall, combining the two terms obtains a better tradeoff between forgetting and intransigence. It indicates preventing over-compact representations while maximizing margins can improve the learned representations that are easier for generalization over previous and new tasks. 
Regarding the weights on the two terms, a larger weight on $\mathcal{L}_{wi}$ is for less compact representations within classes, but a too dispersed representation space may include too much noise. A larger weight on $\mathcal{L}_{bt}$ is more resistant to forgetting but may be less capable of transferring to a new task. 
We show the hyperparameters of all methods in \Cref{tab:hyper}.

\emph{Comparing different replay strategies}: 
We compare \ac{SDRL} using different memory replay strategies (\ac{ER} \& \ac{BER}) in \Cref{tab:drl_rpl}. The results show that \ac{SDRL} has general improvements based on the applied replay strategy. \ac{BER} has consistently shown better performance on forgetting whereas \ac{ER} has shown better performance on intransigence. The performance of \ac{SDRL} obviously correlates to the applied replay strategy, e.g., \ac{SDRL}+\ac{BER} gets better performance on forgetting than \ac{SDRL}+\ac{ER}. We see that on benchmarks which are more difficult in terms of intransigence \ac{SDRL}+\ac{ER} obtains better average accuracy than \ac{SDRL}+\ac{BER}, e.g., on TinyImageNet benchmarks.   

\begin{table*}[th]
\centering 
\caption{Comparing \ac{SDRL} with different memory replay strategies, all criteria are in percentage.}
\label{tab:drl_rpl}
\begin{tabular}{c|c|c|c|c|c}
 \hline
                             &      &    SDRL+ BER              & SDRL + ER & BER & ER                     \\ \hline
    \multirow{3}{*}{P-MNIST} & Avg. Accuracy  & $\mathbf{80.5 \pm 0.4}$    & $78.9 \pm 0.4$        & $79.2 \pm 0.3$        & ${78.2 \pm 0.6}$    \\
                             & Avg. Forgetting & $\mathbf{4.5 \pm 0.2}$   & $6.7 \pm 0.3$         & $5.1 \pm 0.3$         & $7.1 \pm 0.6$       \\
                             & Avg. Intransigence & $2.2 \pm 0.5$ & $\mathbf{1.5 \pm 0.2} $        & ${3.1 \pm 0.2} $        & ${1.9 \pm 0.4}$      \\
                              \hline
    \multirow{3}{*}{S-MNIST} & Avg. Accuracy   & $\mathbf{88.1 \pm 0.6}$   & $84.2 \pm 1.4$        & $85.2 \pm 1.17$        & ${83.2 \pm 1.5}$    \\
                             & Avg. Forgetting & $\mathbf{8.9 \pm 0.7}$   & $16.1 \pm 1.8$        & $13.0 \pm 1.5$        & ${17.2 \pm 1.9}$     \\
                             & Avg. Intransigence & $2.4 \pm 0.3$ & $\mathbf{0.6 \pm 0.1}$         & ${2.1 \pm 0.3}$         & $0.8 \pm 0.1$      \\
                             \hline
    \multirow{3}{*}{Fashion} & Avg. Accuracy  & $\mathbf{77.9 \pm 0.8}$    & $76.5 \pm 1.0$        & $77.0 \pm 0.7$        & ${75.8 \pm 1.4}$    \\
                             & Avg. Forgetting  & $\mathbf{16.6 \pm 1.9}$   & $23.5 \pm 1.9$        & $18.2 \pm 2.7$        & ${24.0 \pm 2.7}$    \\
                             & Avg. Intransigence & $6.5 \pm 1.4$ & $\mathbf{2.5 \pm 1.2}$         & ${6.2 \pm 1.8}$         & $2.7 \pm 1.4$      \\
                              \hline
    \multirow{3}{*}{CIFAR10} & Avg. Accuracy  & $\mathbf{40.4 \pm 1.6}$    & ${36.0 \pm 1.7}$        & $37.3 \pm 1.4$        & ${39.4 \pm 1.7}$    \\
                             & Avg. Forgetting  & $\mathbf{41.4 \pm 3.2}$   & $49.6 \pm 4.6$        & $52.3 \pm 1.6$        & ${50.3 \pm 1.9}$    \\
                             & Avg. Intransigence & ${9.8 \pm 1.5}$ & $7.6 \pm 1.8$         & $4.2 \pm 0.6$         & $\mathbf{3.7 \pm 1.0}$      \\
                             \hline
     \multirow{3}{*}{CIFAR100} & Avg. Accuracy  & $19.3 \pm 0.5$    & $\mathbf{19.6 \pm 1.0}$        & $18.2 \pm 0.4$        & ${18.3 \pm 0.3}$    \\
                             & Avg. Forgetting  & $\mathbf{29.0 \pm 1.0}$   & $34.9 \pm 1.1$        & $35.9 \pm 1.0$        & ${37.0 \pm 1.4}$    \\
                             & Avg. Intransigence & ${9.6 \pm 1.0}$ & $4.0 \pm 0.9$         & ${4.6 \pm 0.6}$         & $\mathbf{3.4 \pm 1.0}$      \\
                             \hline
     \multirow{3}{*}{TinyImageNet} & Avg. Accuracy  & ${8.3 \pm 0.2}$    & $\mathbf{11.3 \pm 1.0}$        & $6.7 \pm 0.8$        & ${7.6 \pm 0.6}$    \\
                             & Avg. Forgetting  & ${27.0 \pm 1.4}$   & $30.0 \pm 0.9$        & $\mathbf{26.0 \pm 1.2}$        & ${28.5 \pm 2.0}$    \\
                             & Avg. Intransigence & ${10.8 \pm 1.5}$ & $\mathbf{5.1 \pm 0.8}$         & ${13.3 \pm 1.9}$         & ${10.1 \pm 2.5}$      \\
                             \hline
    \end{tabular}
\end{table*}

\emph{Comparing different memory budgets}: 
\Cref{fig:msize} compares average accuracy of \acs{SDRL} on MNIST tasks with different memory budgets: fixed total budget vs. fixed budget per task. The fixed total budget (M = 300) gets much better performance than M = 50/task in most tasks of Permuted MNIST and it takes less cost of the memory after task 6. In results of Split MNIST, the fixed total budget (M = 300) gets very similar average accuracy with memory M = 50/class while it takes less cost of the memory after task 3. Since the setting of fixed memory size takes larger memory buffer in early tasks, the results indicate better generalization of early tasks can benefit later tasks, especially for more homogeneous tasks such as Permuted MNIST. However, the fixed memory size may suffer from long task sequence because less samples of each task can be kept in the memory when more tasks have been encountered. The results also align with findings in \citet{chaudhry2019continual} which also suggest a hybrid memory strategy could bring improvements.

\begin{table*}[htb]
    \centering
    \caption{Hyperparameters of all methods}
    \begin{tabular}{lllllll}
    \toprule
                                                                      & P-MNIST     & S-MNIST        & Fashion      & CIFAR-10     & CIFAR-100  & TinyImageNet  \\
                                                                \midrule
    training batch size                                               & 20                 & 10                 & 10                 & 10                 & 10     & 5             \\
    learning rate                                                     & 0.1               & 0.02                & 0.02                & 0.1                & 0.05     & 0.1           \\
    learning rate (A-GEM)                                                    & 0.02               & 0.001                & 0.001                & 0.1                & 0.05     & 0.01           \\
    \begin{tabular}[c]{@{}l@{}}ref batch size\\  (A-GEM)\end{tabular} & 256                & 256                & 256                & 512                & 1500    & 1000            \\
    $\alpha$ of SDRL                                                   & 2                  & 2                  & 2                  & 1                & 1       & 2         \\
    $\lambda$ of SDRL                                                  & $1 \times 10^{-3}$ & $1\times 10^{-2}$  & $1\times 10^{-2}$  & $2\times 10^{-3}$  & $2\times 10^{-3}$  & $5\times 10^{-4}$ \\
    $\lambda$ of Multisim                                             & 5                 & 1                  & 1                  & 2                  & 1       & 5         \\
    $\lambda$ of R-Margin                                             & $2 \times 10^{-5}$ & $1 \times 10^{-3}$ & $1 \times 10^{-3}$ & $1 \times 10^{-4}$ & $1 \times 10^{-3}$ & $1 \times 10^{-3}$
    \\
    $\lambda$ of standalone $\mathcal{L}_{bt}$                                                  & $1 \times 10^{-4}$ & $5\times 10^{-4}$  & $5\times 10^{-4}$  & $2\times 10^{-4}$  & $2\times 10^{-4}$  & $5\times 10^{-5}$
    \\ \bottomrule
    \end{tabular}
    \label{tab:hyper}
\end{table*}

\begin{table}[htb]
    \centering
    \caption{The search range of hyperparameters}
    \begin{tabular}{ll}
    \toprule
                                                                      & The grid-search range       \\
                                                                \midrule
    training batch size                                               & [5,10, 20, 50, 100]                  \\
    learning rate                                                     & [0.001, 0.01, 0.02,0.05, 0.1, 0.2]            \\
    \begin{tabular}[c]{@{}l@{}}ref batch size\\  (A-GEM)\end{tabular} & [128, 256, 512, 1000, 1500, 2000]                \\
    $\alpha$ of SDRL                                                   & [0.1, 0.2, 0.5, 1, 2, 4]       \\
    $\lambda$ of SDRL                                                  & 
    $[1, 2, 5] \times [10^{-5}, 10^{-4}, 10^{-3}, 10^{-2}, 10^{-1}]$\\
    $\lambda$ of Multisim                                             & [10, 8, 6, 5, 4, 3, 2, 1, 0.5, 0.2, 0.1, 0.05]             \\
    $\lambda$ of R-Margin                                             & $[1, 2, 5] \times [10^{-5}, 10^{-4}, 10^{-3}, 10^{-2}, 10^{-1}]$
    \\ \bottomrule
    \end{tabular}
    \label{tab:hyper-range}
    \end{table}

\section{Conclusion}
The two fundamental problems of continual learning with small episodic memories are: (\emph{i}) how to make the best use of episodic memories; and (\emph{ii}) how to construct most representative episodic memories. Gradient-based approaches have demonstrated that the diversity of gradients computed on samples from old and new tasks is a key to generalization over these tasks. 
In this paper we formally connect the diversity of gradients to discriminativeness of representations, which leads to an alternative opportunity to reduce the diversity of gradients in \cl{}.

We subsequently exploit ideas from \acl{DML} for learning more discriminative representations.  In \cl{} we prefer larger margins between classes as in \acl{DML}; however,  \cl{} requires less compact representations for better compatibility with future tasks. Based on these findings, we propose a more efficient approach for a better use of the episodic memory than gradient-based methods can provide. 
Our findings also shed light 
on the question how to construct a better episodic memory: in particular, it is preferable that the memory  preserves general information on non-necessary dimensions as well. 

One important concern in \cl{} is that the model is expected to encounter unknown data distributions. The generalization on known tasks requires more compatibility than static learning and hence brings the question: how can a model learn better representations for unseen tasks? We will leave this question for future work.

\appendices
\section{Proof of Theorems}\label{sec:proof}

\textbf{Notations}: \textbf{\emph{Negative pair}} represents two samples from different classes. \textbf{\emph{Positive pair}} represents two samples from a same class. Let $\mathcal{L}$ represent the softmax cross entropy loss, $\mathbf{W} \in \mathbb{R}^{D \times K} $ is the weight matrix of the linear model, and $\mathbf{x}_n \in \mathbb{R}^{D}$ denotes the input data,  $\mathbf{y}_n \in \mathbb{R}^{K}$ is a one-hot vector that denotes the label of $\mathbf{x}_n$, $D$ is the dimension of representations, 
$K$ is the number of classes. Let $\bm{p}_n = softmax(\mathbf{o}_n)$, where $\mathbf{o}_n = \mathbf{W}^T\mathbf{x}_n$, the gradient $\bm{g}_n = \nabla_{\mathbf{W}} \mathcal{L}(\mathbf{x}_n, \mathbf{y}_n;\mathbf{W})$. $\mathbf{x}_n,\mathbf{x}_m$ are two different samples when $n \neq m$. 
\begin{customlemma}{1}\label{lemma1}
Let $\bm{\epsilon}_n = \bm{p}_{n} - \mathbf{y}_{n}$, we have:
$
       \langle \bm{g}_{n}, \bm{g}_{m} \rangle = \langle \mathbf{x}_{n}, \mathbf{x}_{m} \rangle \langle \bm{\epsilon}_n, \bm{\epsilon}_m \rangle,
$
\end{customlemma}
\begin{proof}
    Let $\bm{\ell}^{'}_n = \partial \mathcal{L}(\mathbf{x}_n,\mathbf{y}_n;\mathbf{W})/\partial \mathbf{o}_n$, by the chain rule, we have $\langle \bm{g}_{n}, \bm{g}_{m} \rangle = \langle \mathbf{x}_{n}, \mathbf{x}_{m} \rangle \langle \bm{\ell}^{'}_{n}, \bm{\ell}^{'}_{m} \rangle$.
    By the definition of softmax cross-entropy loss $\mathcal{L}$, we can find $\bm{\ell}^{'}_n = \bm{p}_{n} - \mathbf{y}_{n} = \bm{\epsilon}_n$. 

\end{proof}

\begin{customthm}{1}\label{corl2}
    Suppose $\mathbf{y}_n \neq \mathbf{y}_m$, and let $c_n$ denote the class index of $\mathbf{x}_n$ (i.e. $\mathbf{y}_{n,c_n} = 1, \mathbf{y}_{n,i}=0, \forall i \neq c_n$). Let $\bm{\beta} \triangleq \bm{p}_{n,c_m} + \bm{p}_{m,c_n}$ and  $\mathbf{s}_p \triangleq \langle \bm{p}_n, \bm{p}_m \rangle$, then:
\begin{equation*}
    \begin{split}
       & \Prob\Bigg(\text{sign}(\langle \bm{g}_{n}, \bm{g}_{m} \rangle)= \text{sign}(-\langle \mathbf{x}_{n}, \mathbf{x}_{m} \rangle)\Bigg) 
        = \Prob(\bm{\beta}  > \mathbf{s}_p),
    \end{split}
\end{equation*}
\end{customthm}
\begin{proof}
    According to Lemma 1 and $\mathbf{y}_n \neq \mathbf{y}_m$, we have 
    \begin{equation*}
        \begin{split}
            &\langle \bm{\epsilon}_{n}, \bm{\epsilon}_{m} \rangle 
            = \langle \bm{p}_{n}, \bm{p}_{m} \rangle - \bm{p}_{n,c_m} - \bm{p}_{m,c_n} = \mathbf{s}_p - \bm{\beta}
        \end{split}
    \end{equation*}
    When $\bm{\beta} > \mathbf{s}_p$, we must have $\langle \bm{\epsilon}_{n}, \bm{\epsilon}_{m} \rangle < 0$. According to Lemma 1, we prove this theorem. 
    \end{proof}

\begin{customthm}{2}\label{corl1}
    Suppose $\mathbf{y}_n = \mathbf{y}_m$, when $\langle \bm{g}_{n}, \bm{g}_{m} \rangle \neq 0$, we have:
    \begin{equation*}
    \text{sign}(\langle \bm{g}_{n}, \bm{g}_{m} \rangle)= \text{sign}(\langle \mathbf{x}_{n}, \mathbf{x}_{m} \rangle)
    \end{equation*}
\end{customthm}
    
    \begin{proof}
        Because $\sum_{k=1}^K \bm{p}_{n,k} = 1$, $ \bm{p}_{n,k} \ge 0, \forall k$, and $c_n = c_m = c$,
        \begin{equation}
            \begin{split}
                \langle \bm{\epsilon}_{n}, \bm{\epsilon}_{m} \rangle = \sum_{k\neq c}^K \bm{p}_{n,k} \bm{p}_{m,k} + (\bm{p}_{n,c}-1)(\bm{p}_{m,c}-1) \ge 0
            \end{split}
        \end{equation}
        According to Lemma 1, we prove the theorem.
        \end{proof}


\bibliographystyle{IEEEtranN}
\bibliography{reference}


\end{document}